\documentclass[runningheads]{llncs}
\usepackage{graphicx}
\usepackage{comment}
\usepackage{amsmath,amssymb} %
\usepackage{color}
\usepackage{hyperref}
\usepackage{breakcites}
\hypersetup{
	colorlinks=true,
	citecolor=blue,
	linkcolor=blue
}

\usepackage[width=122mm,left=12mm,paperwidth=146mm,height=193mm,top=12mm,paperheight=217mm]{geometry}

\usepackage{booktabs} 
\usepackage{pifont}%
\usepackage{enumitem}

\newcommand{\cmark}{\ding{51}}%
\newcommand{\xmark}{\ding{55}}%

\begin{document}
	\pagestyle{headings}
	\mainmatter
	\def\ECCVSubNumber{794}  %
	
	\title{DLow: Diversifying Latent Flows for Diverse Human Motion Prediction} %

	\titlerunning{DLow for Diverse Human Motion Prediction}
	\author{Ye Yuan \and Kris Kitani}
	\authorrunning{Y. Yuan and K. Kitani}
	\institute{Robotics Institute, Carnegie Mellon University\\
		\email{\{yyuan2, kkitani\}@cs.cmu.edu}}
	\maketitle
	\vspace{-4mm}
	
	\begin{abstract}
		Deep generative models are often used for human motion prediction as they are able to model multi-modal data distributions and characterize diverse human behavior. While much care has been taken into designing and learning deep generative models, how to efficiently produce diverse samples from a deep generative model \emph{after} it has been trained is still an under-explored problem. To obtain samples from a pretrained generative model, most existing generative human motion prediction methods draw a set of independent Gaussian latent codes and convert them to motion samples. Clearly, this random sampling strategy is not guaranteed to produce diverse samples for two reasons: (1)~The independent sampling cannot force the samples to be diverse; (2)~The sampling is based solely on likelihood which may only produce samples that correspond to the major modes of the data distribution. To address these problems, we propose a novel sampling method, Diversifying Latent Flows (DLow), to produce a diverse set of samples from a pretrained deep generative model. Unlike random (independent) sampling, the proposed DLow sampling method samples a single random variable and then maps it with a set of learnable mapping functions to a set of correlated latent codes. The correlated latent codes are then decoded into a set of correlated samples. During training, DLow uses a diversity-promoting prior over samples as an objective to optimize the latent mappings to improve sample diversity. The design of the prior is highly flexible and can be customized to generate diverse motions with common features (e.g., similar leg motion but diverse upper-body motion). Our experiments demonstrate that DLow outperforms state-of-the-art baseline methods in terms of sample diversity and accuracy. Our code is released on the project page: \href{https://www.ye-yuan.com/dlow}{\texttt{https://www.ye-yuan.com/dlow}}.
		\vspace{-3mm}
	\end{abstract}
	\vspace{-9mm}
	
	\begin{figure}[ht]
		\centering
		\includegraphics[width=0.9\textwidth]{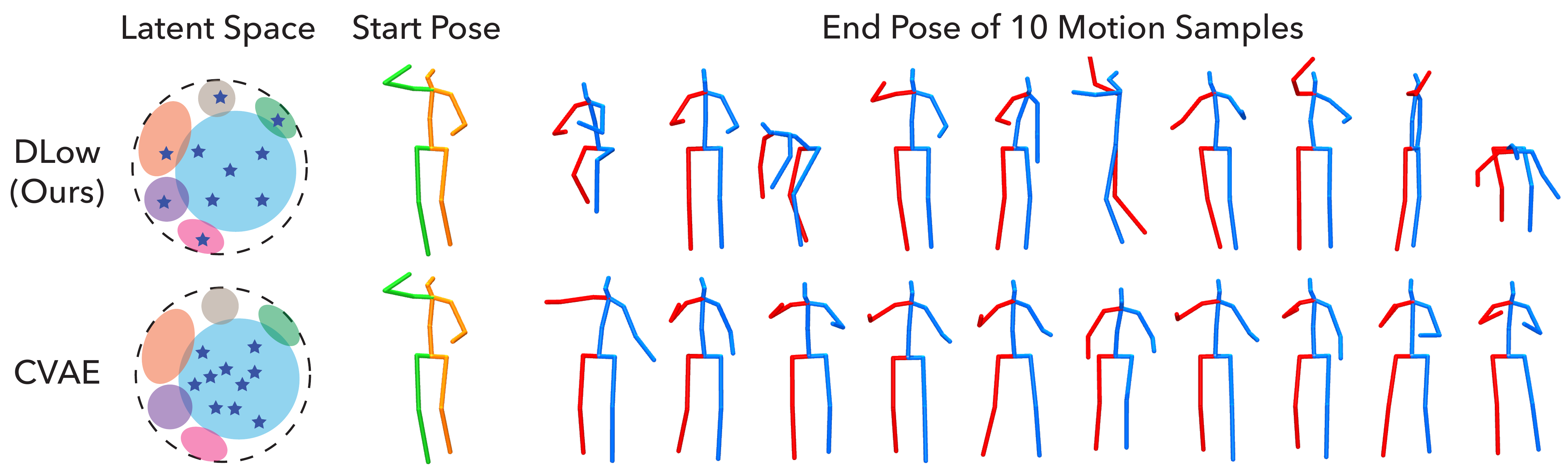}
		\vspace{-4mm}
		\caption{In the latent space of a conditional variational autoencoder (CVAE), samples (stars) from our method DLow are able to cover more modes (colored ellipses) than the CVAE samples. In the motion space, DLow generates a diverse set of future human motions while the CVAE only produces perturbations of the motion of the major mode.}
		\label{fig:teaser}
		\vspace{-7mm}
	\end{figure}

	\section{Introduction}
	\vspace{-1mm}
	
	Human motion prediction, i.e., predicting the future 3D poses of a person based on past poses, is an important problem in computer vision and has many useful applications in autonomous driving~\cite{paden2016survey}, human robot interaction~\cite{koppula2013anticipating} and healthcare~\cite{troje2002decomposing}. It is a challenging problem because the future motion of a person is potentially diverse and multi-modal due to the complex nature of human behavior. For many safety-critical applications, it is important to predict a diverse set of human motions instead of just the most likely one. For examples, an autonomous vehicle should be aware that a nearby pedestrian can suddenly cross the road even though the pedestrian will most likely remain in place. This diversity requirement calls for a generative approach that can fully characterize the multi-modal distribution of future human motion.
	
	Deep generative models, e.g., variational autoencoders (VAEs)~\cite{kingma2013auto}, are effective tools to model multi-modal data distributions. Most existing work~\cite{walker2017pose,lin2018human,barsoum2018hp,ruiz2018human,kundu2019bihmp,yan2018mt,aliakbarian2020stochastic} using deep generative models for human motion prediction is focused on the design of the generative model to allow it to effectively learn the data distribution. After the generative model is learned, little attention has been paid to the sampling method used to produce \emph{motion samples} (predicted future motions) from the \emph{pretrained} generative model (weights kept fixed). Most of prior work predicts a set of motions by randomly sampling a set of latent codes from the latent prior and decoding them with the generator into motion samples. We argue that such a sampling strategy is not guaranteed to produce a diverse set of samples for two reasons: (1) The samples are independently drawn, which makes it difficult to enforce diversity; (2) The samples are drawn based on likelihood only, which means many samples may concentrate around the major modes (which have more observed data) of the data distribution and fail to cover the minor modes (as shown in Fig.~\ref{fig:teaser} (Bottom)). The poor sample efficiency of random sampling means that one needs to draw a large number of samples in order to cover all the modes which is computationally expensive and can lead to high latency, making it unsuitable for real-time applications such as autonomous driving and virtual reality.  This prompts us to address an overlooked aspect of diverse human motion prediction --- the sampling strategy.
	
	We propose a novel sampling method, Diversifying Latent Flows (DLow), to obtain a diverse set of samples from a pretrained deep generative model. For this work, we use a conditional variational autoencoder (CVAE) as our pretrained generative model but other generative models can also be used with our approach. DLow is inspired by the two previously mentioned problems with random (independent) sampling. To tackle problem (1) where sample independence limits model diversity, we introduce a new random variable and a set of learnable deterministic mapping functions to correlate the motion samples. We first transform the random variable with the mappings functions to generate a set of correlated latent codes which are then decoded into motion samples using the generator. As all motion samples are generated from a common random factor, this formulation allows us to model the joint sample distribution and offers us the opportunity to impose diversity on the samples by optimizing the parameters of the mapping functions. To address problem (2) where likelihood-based sampling limits diversity, we introduce a diversity-promoting prior (loss function) on the samples during the training of DLow. The prior follows an energy-based formulation using an energy function based on pairwise sample distance. We optimize the mapping functions during training to minimize the cross entropy between the joint sample distribution and diversity-promoting prior to increase sample diversity. To strike a balance between diversity and likelihood, we add a KL term to the optimization to enhance the likelihood of each sample. The relative weights between the prior term and the KL term represent the trade-off between the diversity and likelihood of the generated motion samples.
	Furthermore, our approach is highly flexible in that by designing different forms of the diversity-promoting prior we can impose a variety of structures on the samples besides diversity. For example, we can design the prior to ask the motion samples to cover the ground truth better to achieve higher sample accuracy. Additionally, other designs of the prior can enable new applications, such as controllable motion prediction, where we generate diverse motion samples that share some common features (e.g., similar leg motion but diverse upper-body motion).
	
	The contributions of this work are the following: 
	(1) We propose a novel perspective for addressing sample diversity in deep generative models --- designing sampling methods for a \emph{pretrained} generative model. 
	(2) We propose a principled sampling method, DLow, which formulates diversity sampling as a constrained optimization problem over a set of learnable mapping functions using a diversity-promoting prior on the samples and KL constraints on the latent codes, which allows us to balance between sample diversity and likelihood.
	(3)~Our approach allows for flexible design of the diversity-promoting prior to obtain more accurate samples or enable new applications such as controllable motion prediction. 
	(4)~We demonstrate through human motion prediction experiments that our approach outperforms state-of-the-art baseline methods in terms of sample diversity and accuracy.

	\vspace{-5mm}
	\section{Related Work}
	\vspace{-2mm}
	\noindent\textbf{Human Motion Prediction.} Most previous work takes a deterministic approach to modeling human motion and regress a single future motion from past 3D poses \cite{fragkiadaki2015recurrent,jain2016structural,butepage2017deep,li2017auto,ghosh2017learning,martinez2017human,pavllo2018quaternet,chiu2019action,gopalakrishnan2019neural,aksan2019structured,wang2019imitation,mao2019learning} or video frames \cite{chao2017forecasting,zhang2019predicting,yuan2019egopose}. While these approaches are able to predict the most likely future motion, they fail to model the multi-modal nature of human motion, which is essential for safety-critical applications. More related to our work, stochastic human motion prediction methods start to gain popularity with the development of deep generative models. These methods~\cite{walker2017pose,lin2018human,barsoum2018hp,ruiz2018human,kundu2019bihmp,yan2018mt,aliakbarian2020stochastic,yuan2020residual} often build upon popular generative models such as conditional generative adversarial networks (CGANs;~\cite{goodfellow2014generative}) or conditional variational autoencoders (CVAEs;~\cite{kingma2013auto}). The aforementioned methods differ in the design of their generative models, but at test time they follow the same sampling strategy --- randomly and independently sampling trajectories from the pretrained generative model without considering the correlation between samples. In this work, we propose a principled sampling method that can produce a diverse set of samples, thus improving sample efficiency compared to the random sampling typically used in prior work.
	
	\vspace{1pt}
	\noindent\textbf{Diverse Inference.} Producing a diverse set of solutions has been investigated in numerous problems in computer vision and machine learning. A branch of these diversity-driven methods stems from the M-Best MAP problem~\cite{nilsson1998efficient,seroussi1994algorithm}, including diverse M-Best solutions~\cite{batra2012diverse} and multiple choice learning~\cite{guzman2012multiple,lee2016stochastic}. Alternatively, submodular function maximization has been applied to select a diverse subset of garments from fashion images~\cite{hsiao2018creating}. Another type of methods~\cite{kulesza2011k,gong2014diverse,gillenwater2014expectation,huang2015we,azadi2017learning,yuan2019diverse,WengYuan2020} seeks diversity using determinantal point processes (DPPs;~\cite{macchi1975coincidence,kulesza2012determinantal}) which are efficient probabilistic models that can measure the global diversity and quality within a set. Similarly, Fisher information~\cite{rissanen1996fisher} has been used for diverse feature~\cite{gu2012generalized} and data~\cite{sourati2017probabilistic} selection. Diversity has also been a key aspect in generative modeling. A vast body of work has tried to alleviate the mode collapse problem in GANs~\cite{che2016mode,chen2016infogan,srivastava2017veegan,arjovsky2017wasserstein,gulrajani2017improved,elfeki2018gdpp,lin2018pacgan,yang2019diversity} and the posterior collapse problem in VAEs~\cite{zhao2017infovae,tolstikhin2017wasserstein,kim2018semi,bhattacharyya2018accurate,liu2019cyclical,he2019lagging}. Normalizing flows~\cite{rezende2015variational} have also been used to promote diversity in trajectory forecasting~\cite{rhinehart2018r2p2,guan2020generative}. This line of work aims to improve the diversity of the data distribution learned by deep generative models. We address diversity from a different angle by improving the strategy for producing samples from a pretrained deep generative model. 
	\vspace{-2mm}
	\section{Diversifying Latent Flows (DLow)}
	\label{sec:dlow}
	\vspace{-5pt}
	
	For many existing methods on generative vision tasks such as multi-modal human motion prediction, the primary focus is to learn a good generative model that can capture the multi-modal distribution of the data. In contrast, once the generative model is learned, little attention has been paid to devising sampling strategies for producing diverse samples from the \emph{pretrained} generative model.
	
	In this section, we will introduce our method, Diversifying Latent Flows (DLow), as a principled way for drawing a diverse and likely set of samples from a pretrained generative model (weights kept fixed). To provide the proper context, we will first start with a brief review of deep generative models and how traditional methods produce samples from a pretrained generative model.
	
	\vspace{2mm}
	\noindent\textbf{Background: Deep Generative Models.} 
	Let $\mathbf{x} \in \mathcal{X}$ denote data (e.g., human motion) drawn from a data distribution $p(\mathbf{x}|\mathbf{c})$ where $\mathbf{c}$ is some conditional information (e.g., past motion). One can reparameterize the data distribution by introducing a latent variable $\mathbf{z} \in \mathcal{Z}$ such that $p(\mathbf{x}|\mathbf{c}) = \int_\mathbf{z} p(\mathbf{x}|\mathbf{z}, \mathbf{c}) p(\mathbf{z})d\mathbf{z}$, where $p(\mathbf{z})$ is a Gaussian prior distribution. Deep generative models learn $p(\mathbf{x}|\mathbf{c})$ by modeling the conditional distribution $p(\mathbf{x}|\mathbf{z}, \mathbf{c})$, and the generative process can be described as sampling $\mathbf{z}$ and mapping them to data samples $\mathbf{x}$ using a deterministic \emph{generator} function $G_\theta: \mathcal{Z} \rightarrow \mathcal{X}$ as
	\begin{align}
	\label{eq:p_z}
	&\mathbf{z} \sim p(\mathbf{z})\,, \\
	\label{eq:G_theta}
	&\mathbf{x} = G_\theta(\mathbf{z}, \mathbf{c})\,,
	\end{align}
	where the generator $G_\theta$ is instantiated as a deep neural network parametrized by $\theta$. This generative process produces samples from the implicit sample distribution $p_\theta(\mathbf{x}|\mathbf{c})$ of the generative model, and the goal of generative modeling is to learn a generator $G_\theta$ such that $p_\theta(\mathbf{x}|\mathbf{c}) \approx p(\mathbf{x}|\mathbf{c})$. There are various approaches for learning the generator function $G_\theta$, which yield different types of deep generative models such as variational autoencoders (VAEs;~\cite{kingma2013auto}), normalizing flows (NFs;~\cite{rezende2015variational}), and generative adversarial networks (GANs;~\cite{goodfellow2014generative}). Note that even though the discussion in this work is focused on conditional generative models, our method can be readily applied to the unconditional case.
	
	\vspace{2mm}
	\noindent\textbf{Random Sampling.} 
	Once the generator function $G_\theta$ is learned, traditional approaches produce samples from the learned data distribution $p_\theta(\mathbf{x}|\mathbf{c})$ by first randomly sampling a set of latent codes $Z = \{\mathbf{z}_1, \ldots, \mathbf{z}_K\}$ from the latent prior $p(\mathbf{z})$ (Eq.~\eqref{eq:p_z}) and decode $Z$ with the generator $G_\theta$ into a set of data samples $X = \{\mathbf{x}_1, \ldots, \mathbf{x}_K\}$ (Eq.~\eqref{eq:G_theta}). We argue that such a sampling strategy may result in a less diverse sample set for two reasons: (1) Independent sampling cannot model the repulsion between samples within a diverse set; (2) The sampling is only based on the data likelihood and many samples can concentrate around a small number of modes that have more training data. As a result, random sampling can lead to low sample efficiency because many samples are similar to one another and fail to cover other modes in the data distribution.
	
	\begin{figure*}[t]
		\centering
		\includegraphics[width=\textwidth]{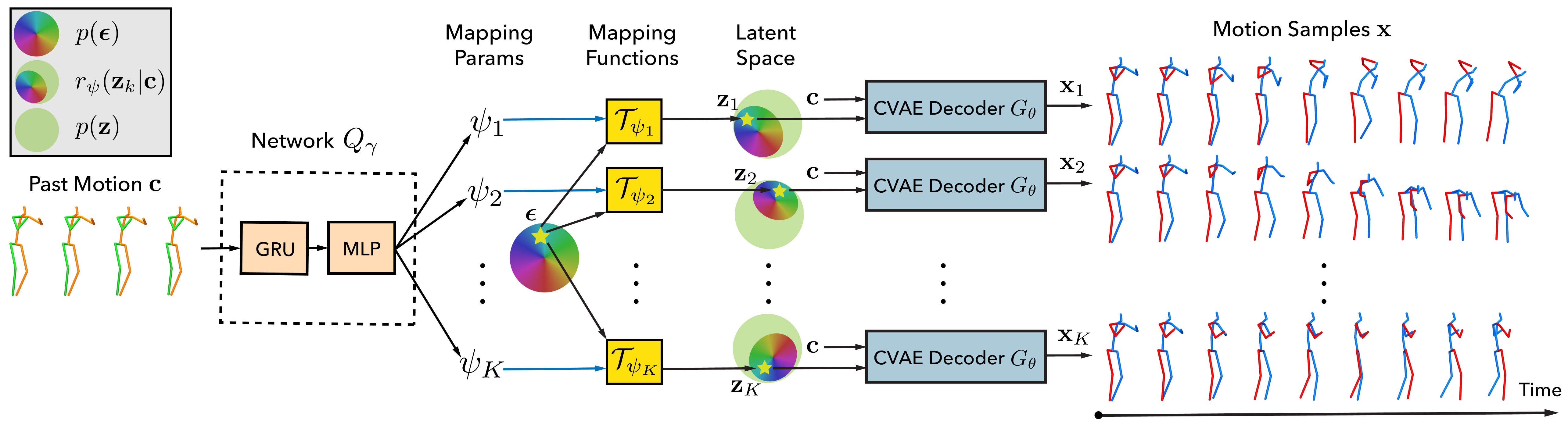}
		\vspace{-7mm}
		\caption{\textbf{Overview of our DLow framework applied to diverse human motion prediction.} The network $Q_\gamma$ takes past motion $\mathbf{c}$ as input and outputs the parameters of the mapping functions $\mathcal{T}_{\psi_1}, \ldots, \mathcal{T}_{\psi_K}$. Each mapping $\mathcal{T}_{\psi_k}$ transforms the random variable $\boldsymbol{\epsilon}$ to a different latent code $\mathbf{z}_k$ and also warps the density $p(\boldsymbol{\epsilon})$ to the latent code density $r_\psi(\mathbf{z}_k|\mathbf{c})$. Each latent code $\mathbf{z}_k$ is decoded by the CVAE decoder into a motion sample $\mathbf{x}_k$.}
		\label{fig:overview}
		\vspace{-5mm}
	\end{figure*}
	
	\vspace{2mm}
	\noindent\textbf{DLow Sampling.} To address the above issues with the random sampling approach, we propose an alternative sampling method, Diversifying Latent Flows (DLow), that can generate a diverse and likely set of samples from a pretrained deep generative model. Again, we stress that the weights of the generative model are kept fixed for DLow. We later apply DLow to the task of human motion prediction in Sec.~\ref{sec:motion_pred} to demonstrate DLow's ability to improve sample diversity.
	
	Instead of sampling each latent code $\mathbf{z}_k \in Z$ independently according to $p(\mathbf{z})$, we introduce a random variable $\boldsymbol{\epsilon}$ and conditionally generate the latent codes $Z$ and data samples $X$ as follows:
	\begin{align}
	& \boldsymbol{\epsilon} \sim p(\boldsymbol{\epsilon})\,, \\
	& \mathbf{z} _k = \mathcal{T}_{\psi_k}(\boldsymbol{\epsilon}) \,,\;\, \quad \quad 1 \leq k \leq K\,, \\
	& \mathbf{x} _k = G_\theta(\mathbf{z}_k, \mathbf{c})\,,	\,\quad 1 \leq k \leq K\,,
	\end{align}
	where $p(\boldsymbol{\epsilon})$ is a Gaussian distribution, $\mathcal{T}_{\psi_1}, \ldots, \mathcal{T}_{\psi_K}$ are latent mapping functions with parameters $\psi = \{ \psi_1, \ldots, \psi_K \}$, and each $\mathcal{T}_{\psi_k}$ maps $\boldsymbol{\epsilon}$ to a different latent code $\mathbf{z}_k$.
	The above generative process defines a joint distribution $r_\psi(X,Z|\mathbf{c})=p_\theta(X|Z, \mathbf{c})r_\psi(Z|\mathbf{c})$ over the samples $X$ and latent codes $Z$, where $p_\theta(X|Z,\mathbf{c})$ is the conditional distribution induced by the generator $G_\theta(\mathbf{z}, \mathbf{c})$. Notice that in our setup, $r_\psi(X,Z|\mathbf{c})$ depends only on $\psi$ as the generator parameters $\theta$ are learned in advance and are kept fixed. The data samples $X$ can be viewed as a sample from the joint sample distribution $r_\psi(X|\mathbf{c})=\int r_\psi(X,Z|\mathbf{c})dZ$ and the latent codes $Z$ can be regarded as a sample from the joint latent distribution $r_\psi(Z|\mathbf{c})$ induced by warping $p(\boldsymbol{\epsilon})$ through $\mathcal{T}_{\psi_1}, \ldots, \mathcal{T}_{\psi_K}$. If we further marginalize out all variables except for $\mathbf{x}_k$ from $r_\psi(X|\mathbf{c})$, we obtain the marginal sample distribution $r_\psi(\mathbf{x}_k|\mathbf{c})$ from which each sample $\mathbf{x}_k$ is drawn. Similarly, each latent code $\mathbf{z}_k \in Z$ can be viewed as a latent sample from the marginal latent distribution $r_\psi(\mathbf{z}_k|\mathbf{c})$.
	
	The above distribution reparametrizations are illustrated in Fig.~\ref{fig:overview}. We can see that all latent codes $Z$ and data samples $X$ are correlated as they are uniquely determined by $\boldsymbol{\epsilon}$, and by sampling $\boldsymbol{\epsilon}$ one can easily produce $Z$ and $X$ from the joint latent distribution $r_\psi(Z|\mathbf{c})$ and joint sample distribution $r_\psi(X|\mathbf{c})$. Because $r_\psi(Z|\mathbf{c})$ and $r_\psi(X|\mathbf{c})$ are controlled by the latent mapping functions $\mathcal{T}_{\psi_1}, \ldots, \mathcal{T}_{\psi_K}$, we can impose structural constraints on $r_\psi(Z|\mathbf{c})$ and $r_\psi(X|\mathbf{c})$ by optimizing the parameters $\psi$ of the latent mapping functions.

	To encourage the diversity of samples $X$, we introduce a diversity-promoting prior $p(X)$ (specific form defined later) and formulate a constrained optimization problem:
	\vspace{-2mm}
	\begin{align}
	\label{eq:obj}
	\min_\psi & \quad -\mathbb{E}_{X \sim r_\psi(X|\mathbf{c})}[\log p(X)]\,, \\
	\label{eq:constraint}
	\text{s.t.} & \quad \text{KL} (r_\psi(\mathbf{z}_k|\mathbf{c}) \| p(\mathbf{z}_k)) = 0\,, \quad 1 \leq k \leq K\,,
	\end{align}
	where we minimize the cross entropy between the sample distribution $r_\psi(X|\mathbf{c})$ and the diversity-promoting prior $p(X)$.
	However, the objective in Eq.~\eqref{eq:obj} alone can result in very low-likelihood samples $\mathbf{x}_k$ corresponding to latent codes $\mathbf{z}_k$ that are far away from the Gaussian prior $p(\mathbf{z}_k)$.
	To ensure that each sample $\mathbf{x}_k$ also has high likelihood under the generative model $p_\theta(\mathbf{x}|\mathbf{c})$,  we add constraints in Eq.~\eqref{eq:constraint} on the KL divergence between $r_\psi(\mathbf{z}_k|\mathbf{c})$ and the Gaussian prior $p(\mathbf{z}_k)$ (same as $p(\mathbf{z})$) to make $r_\psi(\mathbf{z}_k|\mathbf{c}) = p(\mathbf{z}_k)$ and thus $r_\psi(\mathbf{x}_k|\mathbf{c}) = p_\theta(\mathbf{x}_k|\mathbf{c})$ where $r_\psi(\mathbf{x}_k|\mathbf{c})= \int p_\theta(\mathbf{x}_k|\mathbf{z}_k, \mathbf{c})r_\psi(\mathbf{z}_k|\mathbf{c})d\mathbf{z}_k$ and $ p_\theta(\mathbf{x}_k|\mathbf{c})= \int p_\theta(\mathbf{x}_k|\mathbf{z}_k, \mathbf{c})p(\mathbf{z}_k)d\mathbf{z}_k$. To optimize this constrained objective, we soften the constraints with the Lagrangian function:
	\vspace{-2mm}
	\begin{equation}
	\label{eq:dlow_opt}
	\min_\psi \, -\mathbb{E}_{X \sim r_\psi(X|\mathbf{c})}[\log p(X)] +\beta\sum_{k=1}^K\text{KL} (r_\psi(\mathbf{z}_k|\mathbf{c}) \| p(\mathbf{z}_k))\,,
	\end{equation}
	where we use the same Lagrangian multiplier $\beta$ for all constraints. Despite having similar form, the above objective is very \emph{different} from the objective function of $\beta$-VAE~\cite{higgins2017beta} in many ways: (1) our goal is to learn a diverse sampling distribution $r_\psi(X|\mathbf{c})$ for a pretrained generative model rather than learning the generative model itself; (2) The first part in our objective is a diversifying term instead of a reconstruction term; (3) Our objective function applies to most deep generative models, not just VAEs. In this objective, the softening of the hard KL constraints allows for the trade-off between the diversity and likelihood of the samples $X$. For small $\beta$, $r_\psi(\mathbf{z}_k|\mathbf{c})$ is allowed to deviate from $p(\mathbf{z}_k)$ so that $r_\psi(\mathbf{z}_1|\mathbf{c}), \ldots, r_\psi(\mathbf{z}_K|\mathbf{c})$ can potentially attend to different regions in the latent space as shown in Fig.~\ref{fig:overview} (latent space) to further improve sample diversity. For large $\beta$, the objective will focus on minimizing the KL term so that $r_\psi(\mathbf{z}_k|\mathbf{c})\approx p(\mathbf{z}_k)$ and $r_\psi(\mathbf{x}_k|\mathbf{c})\approx p_\theta(\mathbf{x}_k|\mathbf{c})$, and thus the sample $\mathbf{x}_k$ will have high likelihood under $p_\theta(\mathbf{x}_k|\mathbf{c})$.

	The overall DLow objective is defined as:
	\begin{equation}
	\label{eq:dlow_overall}
	L_\text{DLow} = L_{\text{prior}}  + \beta L_{\text{KL}}\,,
	\end{equation}
	where $L_{\text{prior}}$ and $L_{\text{KL}}$ are the first and second term in Eq.~\eqref{eq:dlow_opt} respectively.
	In the following, we will discuss in detail how we design the latent mapping functions $\mathcal{T}_{\psi_1}, \ldots, \mathcal{T}_{\psi_K}$ and the diversity-promoting prior $p(X)$.

	\vspace{2mm}
	\noindent\textbf{Latent Mapping Functions.}
	Each latent mapping $\mathcal{T}_{\psi_k}$ transforms the Gaussian distribution $p(\boldsymbol{\epsilon})$ to the marginal latent distribution $r_\psi(\mathbf{z}_k|\mathbf{c})$ for latent code $\mathbf{z}_k$ where $\mathcal{T}_{\psi_k}$ is also conditioned on $\mathbf{c}$. As $r_\psi(\mathbf{z}_k|\mathbf{c})$ should stay close to the Gaussian latent prior $p(\mathbf{z}_k)$, it would be ideal if the mapping $\mathcal{T}_{\psi_k}$ makes $r_\psi(\mathbf{z}_k|\mathbf{c})$ also a Gaussian. Thus, we design $\mathcal{T}_{\psi_k}$ to be an invertible affine transformation:
	\begin{equation}
	\label{eq:mapping}
	\mathcal{T}_{\psi_k}(\boldsymbol{\epsilon}) = \mathbf{A}_k(\mathbf{c})\boldsymbol{\epsilon} + \mathbf{b}_k(\mathbf{c}) \,,
	\end{equation}
	where the mapping parameters $\psi_k = \{\mathbf{A}_k(\mathbf{c}), \mathbf{b}_k(\mathbf{c})\}$, $\mathbf{A}_k \in \mathbb{R}^{n_z \times n_z}$ is a nonsingular matrix, $\mathbf{b}_k \in \mathbb{R}^{n_z}$ is a vector, and $n_z$ is the number of dimensions for $\mathbf{z}_k$ and $\boldsymbol{\epsilon}$. As shown in Fig.~\ref{fig:overview} and Fig.~\ref{fig:network} (Right), we use a $K$-head network $Q_\gamma(\mathbf{c})$ to output $\psi_1, \ldots, \psi_K$, and the parameters $\gamma$ of the network $Q_\gamma(\mathbf{c})$ are the parameters to be optimized with the DLow objective in Eq.~\eqref{eq:dlow_overall}.
	
	Under the invertible affine transformation $\mathcal{T}_{\psi_k}$, $r_\psi(\mathbf{z}_k|\mathbf{c})$ becomes a Gaussian distribution $\mathcal{N}(\mathbf{b}_k, \mathbf{A}_k\mathbf{A}_k^T)$. This allows us to compute the KL divergence terms in $L_\text{KL}$ analytically:
	\begin{equation}
	\label{eq:dlow_kl}
	\text{KL} (r_\psi(\mathbf{z}_k|\mathbf{c})\| p(\mathbf{z}_k)) = \frac{1}{2}\left(\operatorname{tr}\left(\mathbf{A}_k\mathbf{A}_k^T\right)+\mathbf{b}_k^T\mathbf{b}_k -n_z - \log\det\left(\mathbf{A}_k\mathbf{A}_k^T\right) \right).
	\end{equation}
	The KL divergence is minimized when $r_\psi(\mathbf{z}_k|\mathbf{c}) = p(\mathbf{z}_k)$ which implies that $\mathbf{A}_k\mathbf{A}_k^T = \mathbf{I}$ and $\mathbf{b}_k = \mathbf{0}$. Geometrically, this means that $\mathbf{A}_k$ is in the orthogonal group $O(n_z)$, which includes all rotations and reflections in an $n_z$-dimensional space. This means any mapping $\mathcal{T}_{\psi_k}$ that is a rotation or reflection operation will minimize the KL divergence. As mentioned before, there is a trade-off between diversity and likelihood in Eq.~\eqref{eq:dlow_overall}. To improve sample diversity (minimize $L_\text{prior}$) without compromising likelihood (KL divergence), we can optimize $\mathcal{T}_{\psi_1}, \ldots, \mathcal{T}_{\psi_K}$ to be different rotations or reflections to map $\boldsymbol{\epsilon}$ to different feasible points $\mathbf{z}_1, \ldots, \mathbf{z}_k$ in the latent space. This geometric understanding sheds light on the mapping space admitted by the hard KL constraints. In practice, we use soft KL constraints in the DLow objective to further enlarge the feasible mapping space which allows us to achieve lower $L_\text{prior}$ and better sample diversity.

	\vspace{2mm}
	\noindent\textbf{Diversity-Promoting Prior.} In the DLow objective, a diversity-promoting prior $p(X)$ on the joint sample distribution is used to guide the optimization of the latent mapping functions $\mathcal{T}_{\psi_1}, \ldots, \mathcal{T}_{\psi_K}$.  
	With an energy-based formulation, the prior $p(X)$ can be defined using an energy function $E(X)$:
	\begin{equation}
	\label{eq:dlow_energy}
	p(X) = \exp(-E(X)) / \mathcal{S}\,,
	\end{equation}
	where $\mathcal{S}$ is a normalizing constant.  Dropping the constant $\mathcal{S}$, the first term in Eq.~\eqref{eq:dlow_opt} can be rewritten as
	\begin{equation}
	\label{eq:dlow_diverse}
	L_{\text{prior}} = \mathbb{E}_{X \sim r_\psi(X|\mathbf{c})}[E(X)]\,.
	\end{equation}
	To promote sample diversity of $X$, we design an energy function $E := E_d$ based on a pairwise distance metric $\mathcal{D}$:
	\begin{equation}
	\label{eq:e_diverse}
	E_d(X) = \frac{1}{K(K-1)}\sum_{i=1}^K\sum_{j\neq i}^K \exp\left(-\frac{\mathcal{D}^2(\mathbf{x}_i, \mathbf{x}_j)} {\sigma_d}\right),
	\end{equation}
	where we use the Euclidean distance for $\mathcal{D}$ and an RBF kernel with scale $\sigma_d$. Minimizing $L_{\text{prior}}$ moves the samples towards a lower-energy (diverse) configuration.
	$L_{\text{prior}}$ can be evaluated efficiently with the reparametrization trick~\cite{kingma2013auto}.

	Up to this point, we have described the proposed sampling method, DLow, for generating a diverse set of samples from a pretrained generative model $p_\theta(\mathbf{x}|\mathbf{c})$. By introducing a common random variable $\boldsymbol{\epsilon}$, DLow allows us to generate correlated samples $X$. Moreover, by introducing learnable mapping functions $\mathcal{T}_{\psi_k}$, we can model the joint sample distribution $r_\psi(X|\mathbf{c})$ and impose structural constraints, such as diversity, on the sample set $X$ which cannot be modeled by random sampling from the generative model.

	\section{Diverse Human Motion Prediction}
	\label{sec:motion_pred}
	
	Equipped with a method to generate diverse samples from a pretrained deep generative model, we now turn our attention to the task of diverse human motion prediction.
	Suppose the pose of a person is a $V$-dimensional vector consisting of 3D joint positions, we use $\mathbf{c} \in \mathbb{R}^{H \times V}$ to denote the past motion of $H$ time steps and $\mathbf{x} \in \mathbb{R}^{T \times V}$ to denote the future motion over a future time horizon of $T$. Given a past motion $\mathbf{c}$, the goal of diverse human motion prediction is to generate a diverse set of future motions $X = \{\mathbf{x}_1, \ldots, \mathbf{x}_K\}$.
	
\begin{figure}[t]
		\centering
		\includegraphics[width=\linewidth]{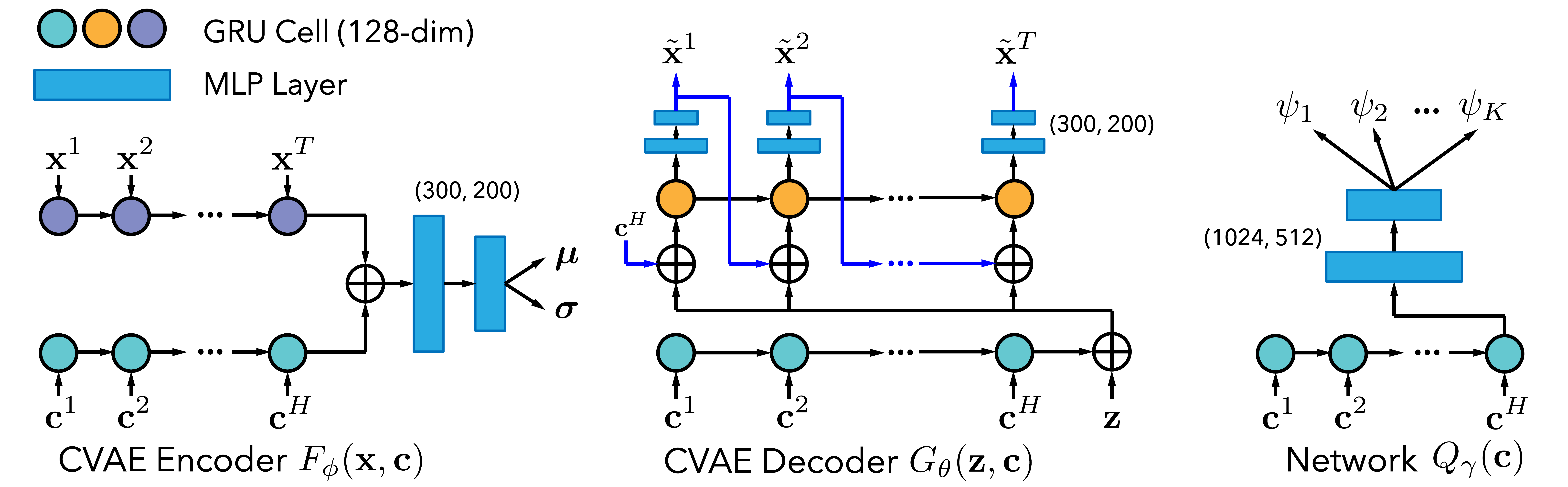}
		\vspace{-7mm}
		\caption{\textbf{Network architectures} for the CVAE and DLow. We use GRUs~\cite{chung2014empirical} to extract motion features. $\mathbf{x}^t$ and $\mathbf{c}^t$ denotes the $t$-th pose in $\mathbf{x}$ and $\mathbf{c}$ respectively.}
		\label{fig:network}
		\vspace{-5mm}
	\end{figure}

	To capture the multi-modal distribution of the future trajectory $\mathbf{x}$, we take a generative approach and use a conditional variational autoencoder (CVAE) to learn the future trajectory distribution $p_\theta(\mathbf{x}|\mathbf{c})$. Here we use the CVAE for its stability over other popular approaches such as CGANs, but other suitable deep generative models could also be used. The CVAE uses a varitional lower bound~\cite{jordan1999introduction} as a surrogate for the intractable true data log-likelihood:
	\begin{equation}
	\label{eq:cvae}
	\mathcal{L}(\mathbf{x} ; \theta, \phi)= \; \mathbb{E}_{q_{\phi}(\mathbf{z} | \mathbf{x}, \mathbf{c})}\left[\log p_{\theta}(\mathbf{x} | \mathbf{z}, \mathbf{c})\right]
	-\operatorname{KL}\left(q_{\phi}(\mathbf{z} | \mathbf{x}, \mathbf{c}) \| p(\mathbf{z})\right),
	\end{equation}
	where $q_{\phi}(\mathbf{z} | \mathbf{x}, \mathbf{c})$ is an $\phi$-parametrized approximate posterior distribution.
	We use multivariate Gaussians for the prior, posterior (encoder distribution) and likelihood (decoder distribution): $p(\mathbf{z})=\mathcal{N}(\mathbf{0}, \mathbf{I})$,  $q_{\phi}(\mathbf{z} | \mathbf{x}, \mathbf{c}) = \mathcal{N}(\boldsymbol{\mu}, \text{Diag}(\boldsymbol{\sigma}^2))$, and $p_\theta(\mathbf{x}|\mathbf{z}, \mathbf{c}) = \mathcal{N}(\tilde{\mathbf{x}}, \alpha\mathbf{I})$ where $\alpha$ is a hyperparameter.
	Both the encoder and decoder are implemented as recurrent neural networks (RNNs). As shown in Fig.~\ref{fig:network}, the encoder network $F_\phi$ outputs the parameters of the posterior distribution: $(\boldsymbol{\mu}, \boldsymbol{\sigma}) = F_\phi(\mathbf{x}, \mathbf{c})$; the decoder network $G_\theta$ outputs the reconstructed future trajectory $\tilde{\mathbf{x}} = G_\theta(\mathbf{z}, \mathbf{c})$. The CVAE is learned via jointly optimizing the encoder and decoder with Eq.~\eqref{eq:cvae}.
	
	\subsection{Diversity Sampling with DLow}
	Once the CVAE is learned, we follow the DLow framework proposed in Sec.~\ref{sec:dlow} to optimize the network $Q_\gamma$ (Fig.~\ref{fig:network} (Right)) and learn the latent mapping functions $\mathcal{T}_{\psi_1}, \ldots, \mathcal{T}_{\psi_K}$. Before doing this, to fully leverage the DLow framework, we will look at one of DLow's key feature, i.e., the design of the diversity-promoting prior $p(X)$ in $L_\text{prior}$ can be flexibly changed by modifying the underlying energy function $E(X)$. This allows us to impose various structural constraints besides diversity on the sample set $X$. Below, we will provide two examples of such prior designs that (1) improve sample accuracy or (2) enable new applications such as controllable motion prediction.
	
	\vspace{1mm}
	\noindent\textbf{Reconstruction Energy.}
	To ensure that the sample set $X$ is both diverse and accurate, i.e., the ground truth future motion $\hat{\mathbf{x}}$ is close to one of the samples in $X$, we can modify the prior's energy function $E$ in Eq.~\eqref{eq:dlow_energy} by adding a reconstruction term $E_r$:
	\begin{align}
	\label{eq:human_prior}
	&E(X) = E_d(X) + \lambda_r E_r(X)\,,\\
	\label{eq:e_recon}
	&E_r(X) = \min_k \mathcal{D}^2(\mathbf{x}_k, \hat{\mathbf{x}})\,,
	\end{align}
	where $\lambda_r$ is a weighting factor and we use Euclidean distance as the distance metric $\mathcal{D}$. As DLow produces a correlated set of samples $X$ instead of independent samples, the network $Q_\gamma$ can learn to distribute samples in a way that are both diverse and accurate, covering the ground truth better. We use this prior design for our main experiments.
	
	\vspace{1mm}
	\noindent\textbf{Controllable Motion Prediction.} 
	Another possible design of the diversity-promoting prior $p(X)$ is one that promotes diversity in a certain subspace of the sample space. In the context of human motion prediction, we may want certain body parts to move similarly but other parts to move differently. For example, we may want leg motion to be similar but upper-body motion to be diverse across motion samples.
	We call this task controllable motion prediction, i.e., finding a set of diverse samples that share some common features, which can allow users or down-stream systems to explore variations of a certain type of samples.
	
	Formally, we divide the human joints into two sets, $J_s$ and $J_d$, and ask samples in $X$ to have similar motions for joints $J_s$ but diverse motions for joints $J_d$. 
	We can slice a motion sample $\mathbf{x}_k$ into two parts: $\mathbf{x}_k = \left(\mathbf{x}_k^s, \mathbf{x}_k^d\right)$ where $\mathbf{x}_k^s$ and $\mathbf{x}_k^d$ correspond to $J_s$ and $J_d$ respectively. Similarly, we can slice the sample set $X$ into two sets: $X_s = \{\mathbf{x}_1^s, \ldots, \mathbf{x}_K^s\}$ and $X_d = \{\mathbf{x}_1^d, \ldots, \mathbf{x}_K^d\}$. We then define a new energy function $E$ for the prior $p(X)$:
	\begin{align}
	\label{eq:human_ctrl}
	&E(X) = E_d(X_d) + \lambda_s E_s(X_s) + \lambda_r E_r(X)\,,\\
	&E_s(X_s) = \frac{1}{K(K-1)}\sum_{i=1}^K\sum_{j\neq i}^K \mathcal{D}^2(\mathbf{x}_i^s, \mathbf{x}_j^s)\,,
	\end{align}
	where we add another energy term $E_s$ weighted by $\lambda_s$ to minimize the motion distance between samples for joints $J_s$, and we only compute the diversity-promoting term $E_d$ using motions of joints $J_d$.
	After optimizing $Q_\gamma$ using the DLow objective with the new energy $E$, we can produce diverse samples $X$ that have similar motions for joints $J_s$.
	
	Furthermore, we may also want to use a reference motion sample $\mathbf{x}_\text{ref}$ to provide the desired features. To achieve this, we can treat $\mathbf{x}_\text{ref}$ as the first sample $\mathbf{x}_1$ in $X$. We first find its corresponding latent code $\mathbf{z}_1 := \mathbf{z}_\text{ref}$ using the CVAE encoder: $\mathbf{z}_\text{ref} =F_\phi^{\boldsymbol{\mu}}(\mathbf{x}_\text{ref}, \mathbf{c})$. We can then find the common variable $\boldsymbol{\epsilon}_\text{ref}$ for generating $X$ using the inverse mapping $\mathcal{T}_{\psi_1}^{-1}$:
	\begin{equation}
	\boldsymbol{\epsilon_\text{ref}} = \mathcal{T}_{\psi_1}^{-1}(\mathbf{z}_\text{ref}) = \mathbf{A}_1^{-1}(\mathbf{z}_\text{ref} - \mathbf{b}_1)\,.
	\end{equation}
	With $\boldsymbol{\epsilon}_\text{ref}$ known, we can generate $X$ that includes $\mathbf{x}_\text{ref}$.
	In practice, we force $\mathcal{T}_{\psi_1}$ to be an identity mapping to enforce $r_\psi(\mathbf{z}_1|\mathbf{c}) = p(\mathbf{z}_1)$ so that $r_\psi(\mathbf{z}_1|\mathbf{c})$ covers the posterior distribution of $\mathbf{z}_\text{ref}$. Otherwise, if $\mathbf{z}_\text{ref}$ lies outside of the high density region of $r_\psi(\mathbf{z}_1|\mathbf{c})$, it may lead to low-likelihood $\boldsymbol{\epsilon}_\text{ref}$ after the inverse mapping.

	\vspace{-3mm}
	\section{Experiments}
	\vspace{-2mm}
	
	\noindent\textbf{Datasets.}
	We perform evaluation on two public motion capture datasets: Human3.6M~\cite{ionescu2013human3} and HumanEva-I~\cite{sigal2010humaneva}. Human3.6M is a large-scale dataset with 11 subjects (7 with ground truth) and 3.6 million video frames in total. Each subject performs 15 actions and the human motion is recorded at 50 Hz. Following previous work \cite{martinez2017simple,luvizon20182d,yang20183d,pavllo20193d}, we adopt a 17-joint skeleton and train on five subjects (S1, S5, S6, S7, S8) and test on two subjects (S9 and S11). HumanEva-I is a relatively small dataset, containing only three subjects recorded at 60 Hz. We adopt a 15-joint skeleton~\cite{pavllo20193d} and use the same train/test split provided in the dataset. By using both a large dataset with more variation in motion and a small dataset with less variation, we can better evaluate the generalization of our method to different types of data. For Human3.6M, we predict future motion for 2 seconds based on observed motion of 0.5 seconds. For HumanEva-I, we forecast future motion for 1 second given observed motion of 0.25 seconds.
	
	\vspace{1mm}
	\noindent\textbf{Baselines.}
	To fully evaluate our method, we consider three types of baselines: (1) Deterministic motion prediction methods, including \textbf{ERD}~\cite{fragkiadaki2015recurrent} and \textbf{acLSTM}~\cite{li2017auto}; (2) Stochastic motion prediction methods, including CVAE based methods, \textbf{Pose-Knows}~\cite{walker2017pose} and \textbf{MT-VAE}~\cite{yan2018mt}, as well as a CGAN based method, \textbf{HP-GAN}~\cite{barsoum2018hp}; (3) Diversity-promoting methods for generative models, including \textbf{Best-of-Many}~\cite{bhattacharyya2018accurate}, \textbf{GMVAE}~\cite{dilokthanakul2016deep}, \textbf{DeLiGAN}~\cite{gurumurthy2017deligan}, and \textbf{DSF}~\cite{yuan2019diverse}.
	
	\vspace{2mm}
	\noindent\textbf{Metrics.} We use the following metrics to measure both sample \emph{diversity} and \emph{accuracy}. (1) \textbf{Average Pairwise Distance (APD)}: average $L2$ distance between all pairs of motion samples to measure diversity within samples, which is computed as $\frac{1}{K(K-1)}\sum_{i=1}^K \sum_{j\neq i}^K \|\mathbf{x}_i - \mathbf{x}_j\|$. (2) \textbf{Average Displacement Error (ADE)}: average $L2$ distance over all time steps between the ground truth motion $\hat{\mathbf{x}}$ and the closest sample, which is computed as $\frac{1}{T}\min_{\mathbf{x} \in X} \|\hat{\mathbf{x}} - \mathbf{x}\|$. (3) \textbf{Final Displacement Error (FDE)}: $L2$ distance between the final ground truth pose $\mathbf{x}^T$ and the closest sample's final pose, which is computed as $\min_{\mathbf{x} \in X} \|\hat{\mathbf{x}}^T - \mathbf{x}^T\|$. (4) \textbf{Multi-Modal ADE (MMADE)}: the multi-modal version of ADE that obtains multi-modal ground truth future motions by grouping similar past motions. (5) \textbf{Multi-Modal FDE (MMFDE)}: the multi-modal version of FDE.
	
	In these metrics, APD has been used to measure sample diversity~\cite{aliakbarian2020stochastic}. ADE and FDE are common metrics for evaluating sample accuracy in trajectory forecasting literature~\cite{alahi2016social,lee2017desire,gupta2018social}. MMADE and MMFDE~\cite{yuan2019diverse} are metrics used to measure a method's ability to produce multi-modal predictions.
	
	\vspace{1mm}
	\noindent\textbf{Implementation Details.} We use a batch size of 64 and set the latent dimensions $n_z$ to 128 in all experiments. For the CVAE, we sample 5000 training examples every epoch and train the networks for 500 epochs using Adam~\cite{kingma2014adam} and a learning rate of~\mbox{1e-3}. The DLow objective in Eq.~\eqref{eq:dlow_overall} can be rewritten as: $L(\psi) = \beta L_\text{KL} + \lambda_d E_d + \lambda_r E_r$. We set $(\beta, \lambda_d, \lambda_r)$ to $(1, 25, 2)$ for Human3.6M and $(1, 50, 2)$ for HumanEva-I. For the mappings $T_{\psi_k}$, we specify $\mathbf{A}_k$ to be diagonal to reduce the output size of $Q_\gamma$. This design is mainly for computational efficiency, as we do find that using a full parametrization of $\mathbf{A}_k$ improves performance. The RBF kernel scale $\sigma_d$ is set to 100 for Human3.6M and 20 for HumanEva-I. For both datasets, we sample 5000 training examples every epoch and train $Q_\gamma$ for 500 epochs using Adam with a learning rate of 1e-4.

	\begin{table}[ht]
		\footnotesize
		\centering
		\resizebox{\columnwidth}{!}{
			\begin{tabular}{@{\hskip 0mm}lcccclcccccl@{\hskip 0mm}}
				\toprule
				& \multicolumn{5}{c}{Human3.6M~\cite{ionescu2013human3}} & & \multicolumn{5}{c}{HumanEva-I~\cite{sigal2010humaneva}} \\ \cmidrule{2-6} \cmidrule{8-12}
				Method & APD $\uparrow$ & ADE $\downarrow$ & FDE $\downarrow$ & MMADE $\downarrow$ & MMFDE $\downarrow$ & & APD $\uparrow$ & ADE $\downarrow$ & FDE $\downarrow$ & MMADE $\downarrow$ & MMFDE $\downarrow$ \\ \midrule
				DLow (Ours) & \textbf{11.741} & \textbf{0.425} & \textbf{0.518} & \textbf{0.495} & \textbf{0.531} &  & \textbf{4.855} & \textbf{0.251} & \textbf{0.268} & \textbf{0.362} & \textbf{0.339} \\
				ERD \cite{fragkiadaki2015recurrent}                     & 0  & 0.722 & 0.969 & 0.776 & 0.995 &  & 0 & 0.382 & 0.461 & 0.521 & 0.595 \\
				acLSTM \cite{li2017auto}                      & 0  & 0.789 & 1.126 & 0.849 & 1.139 &  & 0 & 0.429 & 0.541 & 0.530 & 0.608 \\
				Pose-Knows \cite{walker2017pose}              & 6.723  & 0.461 & 0.560 & 0.522 & 0.569 &  & 2.308 & 0.269 & 0.296 & 0.384 & 0.375 \\
				MT-VAE \cite{yan2018mt}           & 0.403  & 0.457 & 0.595 & 0.716 & 0.883 &  & 0.021 & 0.345 & 0.403 & 0.518 & 0.577 \\
				HP-GAN \cite{barsoum2018hp}           & 7.214  & 0.858 & 0.867 & 0.847 & 0.858 &  & 1.139 & 0.772 & 0.749 & 0.776 & 0.769 \\
				Best-of-Many \cite{bhattacharyya2018accurate} & 6.265  & 0.448 & 0.533 & 0.514 & 0.544 &  & 2.846 & 0.271 & 0.279 & 0.373 & 0.351 \\
				GMVAE \cite{dilokthanakul2016deep}            & 6.769  & 0.461 & 0.555 & 0.524 & 0.566 &  & 2.443 & 0.305 & 0.345 & 0.408 & 0.410 \\
				DeLiGAN \cite{gurumurthy2017deligan}          & 6.509  & 0.483 & 0.534 & 0.520 & 0.545 &  & 2.177 & 0.306 & 0.322 & 0.385 & 0.371 \\
				DSF \cite{yuan2019diverse}                    & 9.330  & 0.493 & 0.592 & 0.550 & 0.599 &  & 4.538 & 0.273 & 0.290 & 0.364 & 0.340 \\
				\bottomrule
			\end{tabular}
		}
		\vspace{1mm}
		\caption{\textbf{Quantitative results} on Human3.6M and HumanEva-I.}
		\label{table:quan}
		\vspace{-5mm}
	\end{table}
	
	\begin{table}[ht]
		\footnotesize
		\centering
		\resizebox{\columnwidth}{!}{
			\begin{tabular}{@{\hskip 1mm}ccc@{\hskip 2mm}cccclcccccl@{\hskip 1mm}}
				\toprule
				\multicolumn{2}{c}{Energy} & & \multicolumn{5}{c}{Human3.6M~\cite{ionescu2013human3}} & & \multicolumn{5}{c}{HumanEva-I~\cite{sigal2010humaneva}} \\ \cmidrule{1-2} \cmidrule{4-8} \cmidrule{10-14}
				$E_d$ & $E_r$& & APD $\uparrow$ & ADE $\downarrow$ & FDE $\downarrow$ & MMADE $\downarrow$ & MMFDE $\downarrow$ & & APD $\uparrow$ & ADE $\downarrow$ & FDE $\downarrow$ & MMADE $\downarrow$ & MMFDE $\downarrow$ \\ \midrule
				\cmark & \cmark & & 11.741 & \textbf{0.425} & \textbf{0.518} & \textbf{0.495} & \textbf{0.531} & & 4.855 & \textbf{0.251} & \textbf{0.268} & \textbf{0.362} & \textbf{0.339}\\  
				\cmark & \xmark & & \textbf{13.091} & 0.546 & 0.663 & 0.599 & 0.669 & & \textbf{4.927} & 0.263 & 0.281 & 0.368  & 0.347\\  
				\xmark & \cmark & & 6.844 & 0.432 & 0.525 & 0.500 & 0.539 & & 2.355 & 0.252 & 0.277 & 0.376 & 0.366 \\  
				\xmark & \xmark & & 6.383 & 0.520 & 0.629 & 0.577 & 0.638 & & 2.247 & 0.281 & 0.317 & 0.395 & 0.393 \\  
				\bottomrule
			\end{tabular}
		}
		\vspace{1mm}
		\caption{\textbf{Ablation study} on Human3.6M and HumanEva-I.}
		\label{table:ablation}
		\vspace{-10mm}
	\end{table}
	
	\vspace{-3mm}
	\subsection{Quantitative Results}
	\vspace{-1mm}
	We summarize the quantitative results on Human3.6M and HumanEva-I in Table~\ref{table:quan}. The metrics are computed with the sample set size $K = 50$. For both datasets, we can see that our method, DLow, outperforms all baselines in terms of both sample diversity (APD) and accuracy (ADE, FDE) as well as covering multi-modal ground truth (MMADE, MMFDE). Determinstic methods like ERD~\cite{fragkiadaki2015recurrent} and acLSTM~\cite{li2017auto} do not perform well because they only predict one future trajectory which can lead to mode averaging. Methods like MT-VAE~\cite{yan2018mt} produce trajectories samples that lack diversity so they fail to cover the multi-modal ground-truth (indicated by high MMADE and MMFDE) despite having decently low ADE and FDE. We would also like to point out the closest competitor DSF~\cite{yuan2019diverse} can only generate one deterministic set of samples, while our method can produce multiple diverse sets by sampling $\boldsymbol{\epsilon}$. We also show how each metric changes against various $K$ in Appendix~\ref{sec:metrics_vs_k}.
	
	\vspace{1mm}
	\noindent\textbf{Ablation Study.}
	We further perform an ablation study (Table~\ref{table:ablation}) to analyze the effects of the two energy terms $E_d$ and $E_r$ in Eq.~\eqref{eq:human_prior}. First, without the reconstruction term $E_r$, the DLow variant is able to achieve higher diversity (APD) at the cost of sample accuracy (ADE, FDE, MMADE, MMFDE). This is expected because the network only optimizes the diversity term $E_d$ and focuses solely on diversity. Second, for the variant without $E_d$, both sample diversity and accuracy decrease. It is intuitive to see why the diversity (APD) decreases. To see why the sample accuracy (ADE, FDE, MMADE, MMFDE) also decreases, we should consider the fact that a more diverse set of samples have a better chance at covering the ground truth. Finally, when we remove both $E_d$ and $E_r$ (i.e., only optimize $L_\text{KL}$), the results are the worst, which is expected.
	
	\begin{figure}[t]
		\centering
		\includegraphics[width=\linewidth]{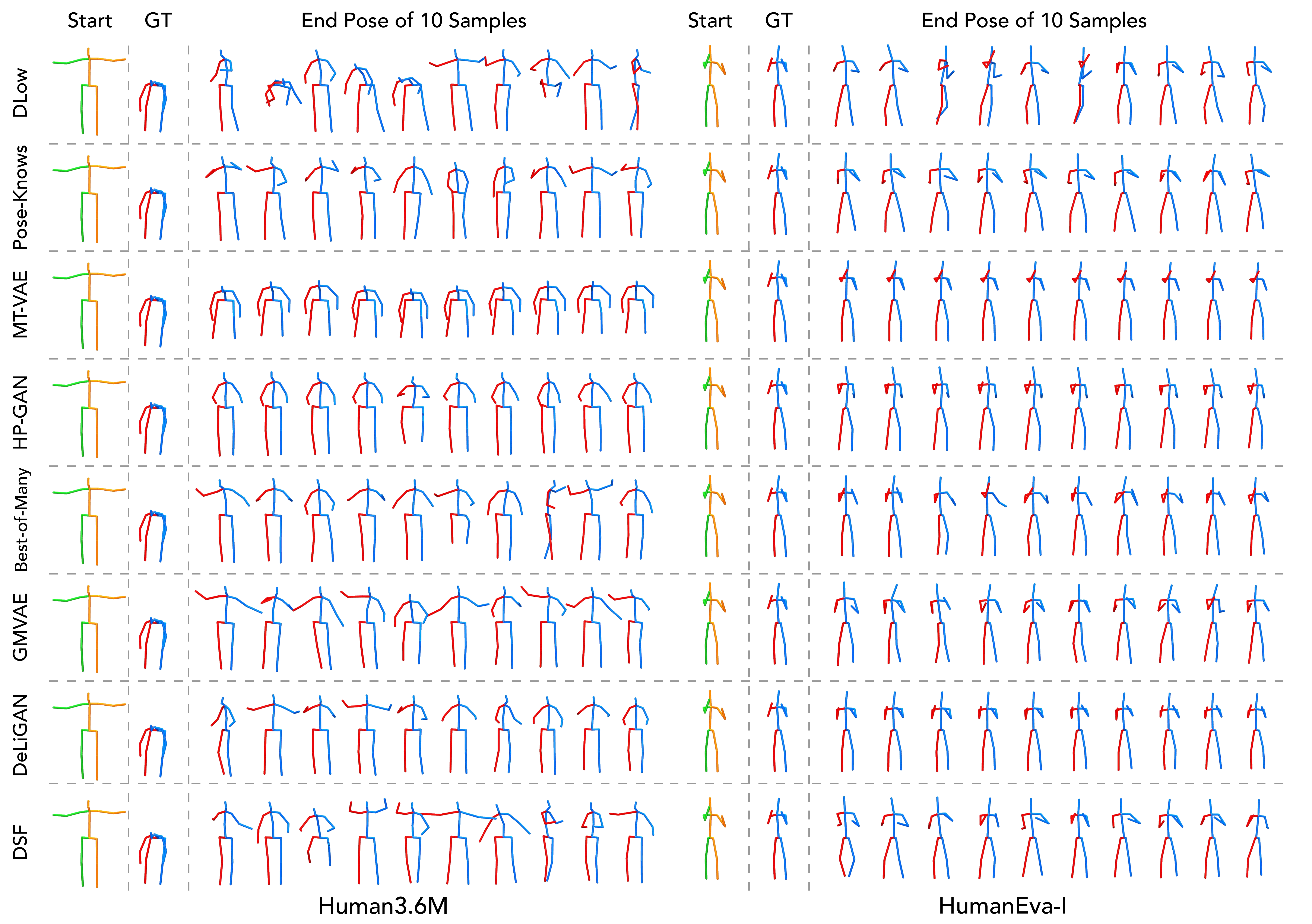}
		\vspace{-7mm}
		\caption{\textbf{Qualitative Results} on Human3.6M and HumanEva-I.}
		\label{fig:comp_base}
		\vspace{-3mm}
	\end{figure}

	\begin{figure}[t]
		\centering
		\includegraphics[width=\linewidth]{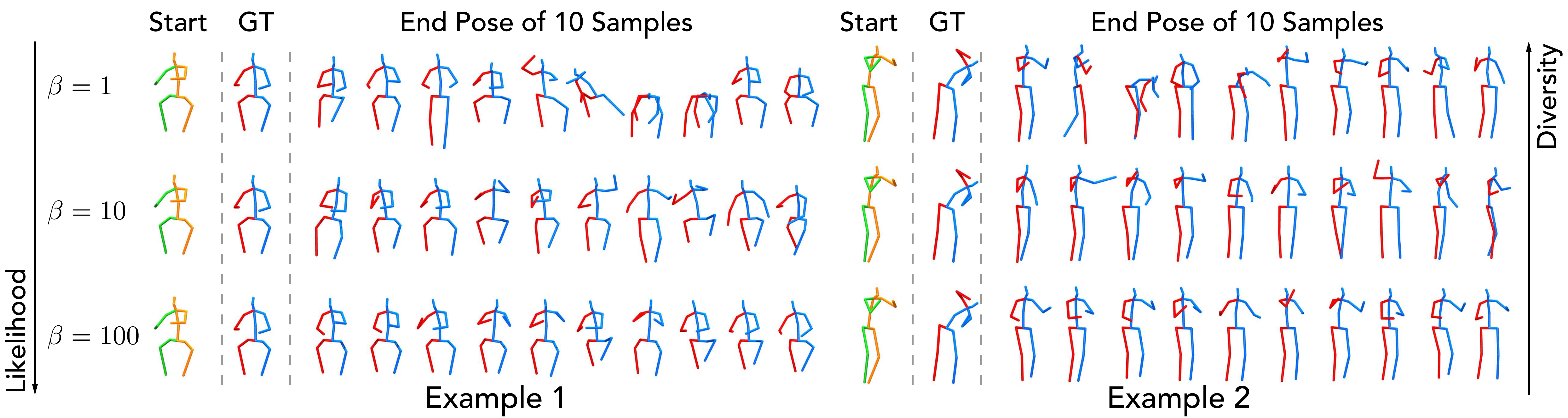}
		\vspace{-7mm}
		\caption{\textbf{Varying $\beta$ in DLow} allows us to balance between diversity and likelihood. }
		\label{fig:beta}
		\vspace{-5mm}
	\end{figure}

	\vspace{-3mm}
	\subsection{Qualitative Results}
	\vspace{-1mm}
	To visually evaluate the diversity and accuracy of each method, we present a qualitative comparison in Fig.~\ref{fig:comp_base} where we render the start pose, the end pose of the ground truth future motion, and the end pose of 10 motion samples. Note that we do not model the global translation of the person, which is why some sitting motions appear to be floating. For Human3.6M, we can see that our method DLow can predict a wide array of future motions, including standing, sitting, bending, crouching, and turning, which cover the ground truth bending motion. In contrast, the baseline methods mostly produce perturbations of a single motion --- standing. For HumanEva-I, we can see that DLow produces interesting variations of the fighting motion, while the baselines produce almost identical future motions.
	
	\begin{figure}[t]
		\centering
		\includegraphics[width=0.9\linewidth]{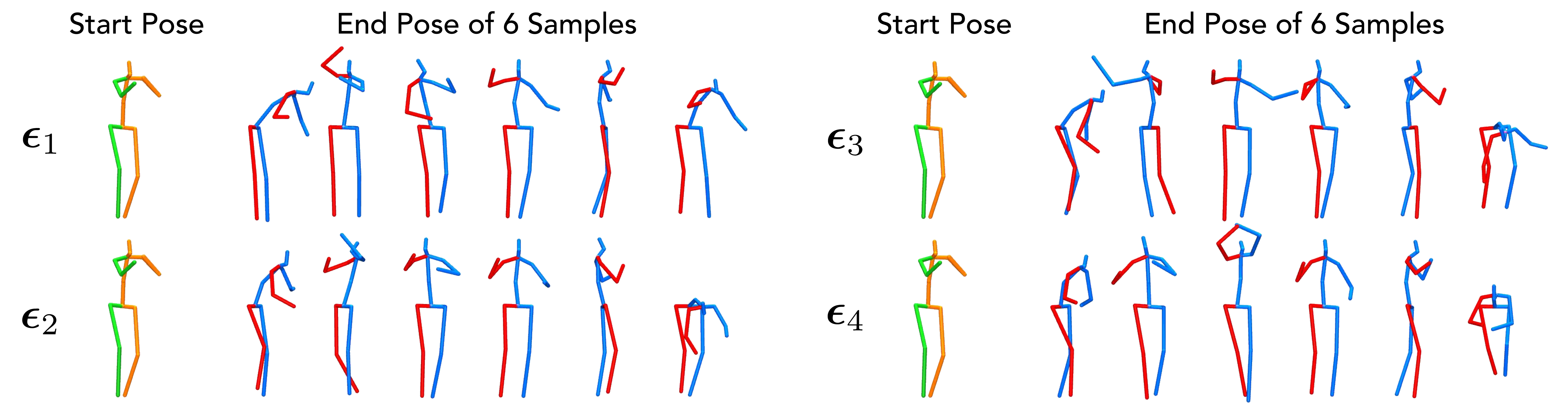}
		\vspace{-5mm}
		\caption{\textbf{Effect of varying $\boldsymbol{\epsilon}$} on motion samples. }
		\label{fig:var}
		\vspace{-3mm}
	\end{figure}
	
	\vspace{1mm}
	\noindent\textbf{Diversity vs. Likelihood.} As discussed in the approach section, the $\beta$ in Eq.~\eqref{eq:dlow_opt} represents the trade-off between sample diversity and likelihood. To verify this, we trained three DLow models with different $\beta$ (1, 10, 100) and visualize the motion samples generated by each model in Fig.~\ref{fig:beta}. We can see that a larger $\beta$ leads to less diverse samples which correspond to the major mode of the generator distribution, while a smaller $\beta$ can produce more diverse motion samples covering other plausible yet less likely future motions.
	
	\vspace{1mm}
	\noindent\textbf{Effect of varying $\boldsymbol{\epsilon}$.} A key difference between our method and DSF~\cite{yuan2019diverse} is that we can generate multiple diverse sets of samples while DSF can only produce a fixed diverse set. To demonstrate this, we show in Fig.~\ref{fig:var} how the motion samples of DLow change with different $\boldsymbol{\epsilon}$. By comparing the four sets of motion samples, one can conclude that changing $\boldsymbol{\epsilon}$ varies each set of samples but preserves the main structure of each motion.
	
\begin{figure}[t]
		\centering
		\includegraphics[width=\linewidth]{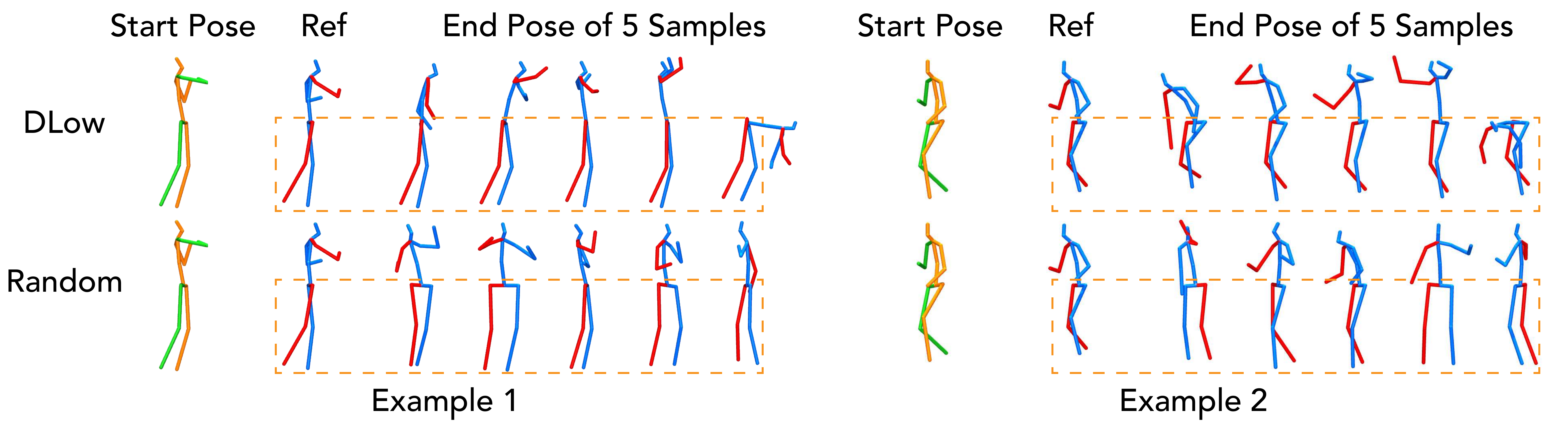}
		\vspace{-7mm}
		\caption{\textbf{Controllable Motion Prediction.} DLow enables samples to have more similar leg motion to the reference.}
		\label{fig:control}
		\vspace{-6mm}
	\end{figure}

	\vspace{1mm}
	\noindent\textbf{Controllable Motion Prediction.}
	As highlighted before, the flexible design of the diversity-promoting prior enables a new application, controllable motion prediction, where we predict diverse motions that share some common features. We showcase this application by conducting an experiment using the energy function defined in Eq.~\eqref{eq:human_ctrl}. 
	The network is trained so that the leg motion of the motion samples is similar while the upper-body motion is diverse.
	The results are shown in Fig.~\ref{fig:control}. We can see that given a reference motion, our method can generate diverse upper-body motion and preserve similar leg motion, while random samples from the CVAE cannot enforce similar leg motion. Please refer to Appendix~\ref{sec:control_res} for more results.

	\vspace{-2mm}
	\section{Conclusion}
	\vspace{-1mm}
	We have proposed a novel sampling strategy, DLow, for deep generative models to obtain a diverse set of future human motions. We introduced learnable latent mapping functions which allowed us to generate a set of correlated samples, whose diversity can be optimized by a diversity-promoting prior. Experiments demonstrated superior performance in generating diverse motion samples. Moreover, we showed that the flexible design of the diversity-promoting prior further enables new applications, such as controllable human motion prediction. We hope that our exploration of deep generative models through the lens of diversity will encourage more work towards understanding the complex nature of modeling and predicting future human behavior.

	\bibliographystyle{splncs04}
	\bibliography{main}

\begin{thebibliography}{10}
\providecommand{\url}[1]{\texttt{#1}}
\providecommand{\urlprefix}{URL }
\providecommand{\doi}[1]{https://doi.org/#1}

\bibitem{aksan2019structured}
Aksan, E., Kaufmann, M., Hilliges, O.: Structured prediction helps 3d human
  motion modelling. In: Proceedings of the IEEE International Conference on
  Computer Vision. pp. 7144--7153 (2019)

\bibitem{alahi2016social}
Alahi, A., Goel, K., Ramanathan, V., Robicquet, A., Fei-Fei, L., Savarese, S.:
  Social lstm: Human trajectory prediction in crowded spaces. In: Proceedings
  of the IEEE Conference on Computer Vision and Pattern Recognition. pp.
  961--971 (2016)

\bibitem{aliakbarian2020stochastic}
Aliakbarian, S., Saleh, F.S., Salzmann, M., Petersson, L., Gould, S.: A
  stochastic conditioning scheme for diverse human motion prediction. In:
  Proceedings of the IEEE/CVF Conference on Computer Vision and Pattern
  Recognition. pp. 5223--5232 (2020)

\bibitem{arjovsky2017wasserstein}
Arjovsky, M., Chintala, S., Bottou, L.: Wasserstein gan. arXiv preprint
  arXiv:1701.07875  (2017)

\bibitem{azadi2017learning}
Azadi, S., Feng, J., Darrell, T.: Learning detection with diverse proposals.
  In: Proceedings of the IEEE Conference on Computer Vision and Pattern
  Recognition. pp. 7149--7157 (2017)

\bibitem{barsoum2018hp}
Barsoum, E., Kender, J., Liu, Z.: Hp-gan: Probabilistic 3d human motion
  prediction via gan. In: Proceedings of the IEEE Conference on Computer Vision
  and Pattern Recognition Workshops. pp. 1418--1427 (2018)

\bibitem{batra2012diverse}
Batra, D., Yadollahpour, P., Guzman-Rivera, A., Shakhnarovich, G.: Diverse
  m-best solutions in markov random fields. In: European Conference on Computer
  Vision. pp. 1--16. Springer (2012)

\bibitem{bhattacharyya2018accurate}
Bhattacharyya, A., Schiele, B., Fritz, M.: Accurate and diverse sampling of
  sequences based on a “best of many” sample objective. In: Proceedings of
  the IEEE Conference on Computer Vision and Pattern Recognition. pp.
  8485--8493 (2018)

\bibitem{butepage2017deep}
Butepage, J., Black, M.J., Kragic, D., Kjellstrom, H.: Deep representation
  learning for human motion prediction and classification. In: Proceedings of
  the IEEE Conference on Computer Vision and Pattern Recognition. pp.
  6158--6166 (2017)

\bibitem{chao2017forecasting}
Chao, Y.W., Yang, J., Price, B., Cohen, S., Deng, J.: Forecasting human
  dynamics from static images. In: Proceedings of the IEEE Conference on
  Computer Vision and Pattern Recognition. pp. 548--556 (2017)

\bibitem{che2016mode}
Che, T., Li, Y., Jacob, A.P., Bengio, Y., Li, W.: Mode regularized generative
  adversarial networks. arXiv preprint arXiv:1612.02136  (2016)

\bibitem{chen2016infogan}
Chen, X., Duan, Y., Houthooft, R., Schulman, J., Sutskever, I., Abbeel, P.:
  Infogan: Interpretable representation learning by information maximizing
  generative adversarial nets. In: Advances in neural information processing
  systems. pp. 2172--2180 (2016)

\bibitem{chiu2019action}
Chiu, H.k., Adeli, E., Wang, B., Huang, D.A., Niebles, J.C.: Action-agnostic
  human pose forecasting. In: 2019 IEEE Winter Conference on Applications of
  Computer Vision (WACV). pp. 1423--1432. IEEE (2019)

\bibitem{chung2014empirical}
Chung, J., Gulcehre, C., Cho, K., Bengio, Y.: Empirical evaluation of gated
  recurrent neural networks on sequence modeling. arXiv preprint
  arXiv:1412.3555  (2014)

\bibitem{dilokthanakul2016deep}
Dilokthanakul, N., Mediano, P.A., Garnelo, M., Lee, M.C., Salimbeni, H.,
  Arulkumaran, K., Shanahan, M.: Deep unsupervised clustering with gaussian
  mixture variational autoencoders. arXiv preprint arXiv:1611.02648  (2016)

\bibitem{elfeki2018gdpp}
Elfeki, M., Couprie, C., Riviere, M., Elhoseiny, M.: Gdpp: Learning diverse
  generations using determinantal point process. arXiv preprint
  arXiv:1812.00068  (2018)

\bibitem{fragkiadaki2015recurrent}
Fragkiadaki, K., Levine, S., Felsen, P., Malik, J.: Recurrent network models
  for human dynamics. In: Proceedings of the IEEE International Conference on
  Computer Vision. pp. 4346--4354 (2015)

\bibitem{ghosh2017learning}
Ghosh, P., Song, J., Aksan, E., Hilliges, O.: Learning human motion models for
  long-term predictions. In: 2017 International Conference on 3D Vision (3DV).
  pp. 458--466. IEEE (2017)

\bibitem{gillenwater2014expectation}
Gillenwater, J.A., Kulesza, A., Fox, E., Taskar, B.: Expectation-maximization
  for learning determinantal point processes. In: Advances in Neural
  Information Processing Systems. pp. 3149--3157 (2014)

\bibitem{gong2014diverse}
Gong, B., Chao, W.L., Grauman, K., Sha, F.: Diverse sequential subset selection
  for supervised video summarization. In: Advances in Neural Information
  Processing Systems. pp. 2069--2077 (2014)

\bibitem{goodfellow2014generative}
Goodfellow, I., Pouget-Abadie, J., Mirza, M., Xu, B., Warde-Farley, D., Ozair,
  S., Courville, A., Bengio, Y.: Generative adversarial nets. In: Advances in
  neural information processing systems. pp. 2672--2680 (2014)

\bibitem{gopalakrishnan2019neural}
Gopalakrishnan, A., Mali, A., Kifer, D., Giles, L., Ororbia, A.G.: A neural
  temporal model for human motion prediction. In: Proceedings of the IEEE
  Conference on Computer Vision and Pattern Recognition. pp. 12116--12125
  (2019)

\bibitem{gu2012generalized}
Gu, Q., Li, Z., Han, J.: Generalized fisher score for feature selection. arXiv
  preprint arXiv:1202.3725  (2012)

\bibitem{guan2020generative}
Guan, J., Yuan, Y., Kitani, K.M., Rhinehart, N.: Generative hybrid
  representations for activity forecasting with no-regret learning. In:
  Proceedings of the IEEE Conference on Computer Vision and Pattern Recognition
  (2020)

\bibitem{gulrajani2017improved}
Gulrajani, I., Ahmed, F., Arjovsky, M., Dumoulin, V., Courville, A.C.: Improved
  training of wasserstein gans. In: Advances in neural information processing
  systems. pp. 5767--5777 (2017)

\bibitem{gupta2018social}
Gupta, A., Johnson, J., Fei-Fei, L., Savarese, S., Alahi, A.: Social gan:
  Socially acceptable trajectories with generative adversarial networks. In:
  Proceedings of the IEEE Conference on Computer Vision and Pattern
  Recognition. pp. 2255--2264 (2018)

\bibitem{gurumurthy2017deligan}
Gurumurthy, S., Kiran~Sarvadevabhatla, R., Venkatesh~Babu, R.: Deligan:
  Generative adversarial networks for diverse and limited data. In: Proceedings
  of the IEEE Conference on Computer Vision and Pattern Recognition. pp.
  166--174 (2017)

\bibitem{guzman2012multiple}
Guzman-Rivera, A., Batra, D., Kohli, P.: Multiple choice learning: Learning to
  produce multiple structured outputs. In: Advances in Neural Information
  Processing Systems. pp. 1799--1807 (2012)

\bibitem{he2019lagging}
He, J., Spokoyny, D., Neubig, G., Berg-Kirkpatrick, T.: Lagging inference
  networks and posterior collapse in variational autoencoders. arXiv preprint
  arXiv:1901.05534  (2019)

\bibitem{higgins2017beta}
Higgins, I., Matthey, L., Pal, A., Burgess, C., Glorot, X., Botvinick, M.,
  Mohamed, S., Lerchner, A.: beta-vae: Learning basic visual concepts with a
  constrained variational framework. Iclr  \textbf{2}(5), ~6 (2017)

\bibitem{hsiao2018creating}
Hsiao, W.L., Grauman, K.: Creating capsule wardrobes from fashion images. In:
  Proceedings of the IEEE Conference on Computer Vision and Pattern
  Recognition. pp. 7161--7170 (2018)

\bibitem{huang2015we}
Huang, D.A., Ma, M., Ma, W.C., Kitani, K.M.: How do we use our hands?
  discovering a diverse set of common grasps. In: Proceedings of the IEEE
  Conference on Computer Vision and Pattern Recognition. pp. 666--675 (2015)

\bibitem{ionescu2013human3}
Ionescu, C., Papava, D., Olaru, V., Sminchisescu, C.: Human3. 6m: Large scale
  datasets and predictive methods for 3d human sensing in natural environments.
  IEEE transactions on pattern analysis and machine intelligence
  \textbf{36}(7),  1325--1339 (2013)

\bibitem{jain2016structural}
Jain, A., Zamir, A.R., Savarese, S., Saxena, A.: Structural-rnn: Deep learning
  on spatio-temporal graphs. In: Proceedings of the IEEE Conference on Computer
  Vision and Pattern Recognition. pp. 5308--5317 (2016)

\bibitem{jordan1999introduction}
Jordan, M.I., Ghahramani, Z., Jaakkola, T.S., Saul, L.K.: An introduction to
  variational methods for graphical models. Machine learning  \textbf{37}(2),
  183--233 (1999)

\bibitem{kim2018semi}
Kim, Y., Wiseman, S., Miller, A.C., Sontag, D., Rush, A.M.: Semi-amortized
  variational autoencoders. arXiv preprint arXiv:1802.02550  (2018)

\bibitem{kingma2014adam}
Kingma, D.P., Ba, J.: Adam: A method for stochastic optimization. arXiv
  preprint arXiv:1412.6980  (2014)

\bibitem{kingma2013auto}
Kingma, D.P., Welling, M.: Auto-encoding variational bayes. arXiv preprint
  arXiv:1312.6114  (2013)

\bibitem{koppula2013anticipating}
Koppula, H.S., Saxena, A.: Anticipating human activities for reactive robotic
  response. In: IROS. p.~2071. Tokyo (2013)

\bibitem{kulesza2011k}
Kulesza, A., Taskar, B.: k-dpps: Fixed-size determinantal point processes. In:
  Proceedings of the 28th International Conference on Machine Learning
  (ICML-11). pp. 1193--1200 (2011)

\bibitem{kulesza2012determinantal}
Kulesza, A., Taskar, B., et~al.: Determinantal point processes for machine
  learning. Foundations and Trends{\textregistered} in Machine Learning
  \textbf{5}(2--3),  123--286 (2012)

\bibitem{kundu2019bihmp}
Kundu, J.N., Gor, M., Babu, R.V.: Bihmp-gan: bidirectional 3d human motion
  prediction gan. In: Proceedings of the AAAI Conference on Artificial
  Intelligence. vol.~33, pp. 8553--8560 (2019)

\bibitem{lee2017desire}
Lee, N., Choi, W., Vernaza, P., Choy, C.B., Torr, P.H., Chandraker, M.: Desire:
  Distant future prediction in dynamic scenes with interacting agents. In:
  Proceedings of the IEEE Conference on Computer Vision and Pattern
  Recognition. pp. 336--345 (2017)

\bibitem{lee2016stochastic}
Lee, S., Prakash, S.P.S., Cogswell, M., Ranjan, V., Crandall, D., Batra, D.:
  Stochastic multiple choice learning for training diverse deep ensembles. In:
  Advances in Neural Information Processing Systems. pp. 2119--2127 (2016)

\bibitem{li2017auto}
Li, Z., Zhou, Y., Xiao, S., He, C., Huang, Z., Li, H.: Auto-conditioned
  recurrent networks for extended complex human motion synthesis. arXiv
  preprint arXiv:1707.05363  (2017)

\bibitem{lin2018human}
Lin, X., Amer, M.R.: Human motion modeling using dvgans. arXiv preprint
  arXiv:1804.10652  (2018)

\bibitem{lin2018pacgan}
Lin, Z., Khetan, A., Fanti, G., Oh, S.: Pacgan: The power of two samples in
  generative adversarial networks. In: Advances in Neural Information
  Processing Systems. pp. 1498--1507 (2018)

\bibitem{liu2019cyclical}
Liu, X., Gao, J., Celikyilmaz, A., Carin, L., et~al.: Cyclical annealing
  schedule: A simple approach to mitigating kl vanishing. arXiv preprint
  arXiv:1903.10145  (2019)

\bibitem{luvizon20182d}
Luvizon, D.C., Picard, D., Tabia, H.: 2d/3d pose estimation and action
  recognition using multitask deep learning. In: Proceedings of the IEEE
  Conference on Computer Vision and Pattern Recognition. pp. 5137--5146 (2018)

\bibitem{macchi1975coincidence}
Macchi, O.: The coincidence approach to stochastic point processes. Advances in
  Applied Probability  \textbf{7}(1),  83--122 (1975)

\bibitem{mao2019learning}
Mao, W., Liu, M., Salzmann, M., Li, H.: Learning trajectory dependencies for
  human motion prediction. In: Proceedings of the IEEE International Conference
  on Computer Vision. pp. 9489--9497 (2019)

\bibitem{martinez2017human}
Martinez, J., Black, M.J., Romero, J.: On human motion prediction using
  recurrent neural networks. In: Proceedings of the IEEE Conference on Computer
  Vision and Pattern Recognition. pp. 2891--2900 (2017)

\bibitem{martinez2017simple}
Martinez, J., Hossain, R., Romero, J., Little, J.J.: A simple yet effective
  baseline for 3d human pose estimation. In: Proceedings of the IEEE
  International Conference on Computer Vision. pp. 2640--2649 (2017)

\bibitem{nilsson1998efficient}
Nilsson, D.: An efficient algorithm for finding the m most probable
  configurationsin probabilistic expert systems. Statistics and computing
  \textbf{8}(2),  159--173 (1998)

\bibitem{paden2016survey}
Paden, B., {\v{C}}{\'a}p, M., Yong, S.Z., Yershov, D., Frazzoli, E.: A survey
  of motion planning and control techniques for self-driving urban vehicles.
  IEEE Transactions on intelligent vehicles  \textbf{1}(1),  33--55 (2016)

\bibitem{pavllo20193d}
Pavllo, D., Feichtenhofer, C., Grangier, D., Auli, M.: 3d human pose estimation
  in video with temporal convolutions and semi-supervised training. In:
  Proceedings of the IEEE Conference on Computer Vision and Pattern
  Recognition. pp. 7753--7762 (2019)

\bibitem{pavllo2018quaternet}
Pavllo, D., Grangier, D., Auli, M.: Quaternet: A quaternion-based recurrent
  model for human motion. arXiv preprint arXiv:1805.06485  (2018)

\bibitem{rezende2015variational}
Rezende, D.J., Mohamed, S.: Variational inference with normalizing flows. arXiv
  preprint arXiv:1505.05770  (2015)

\bibitem{rhinehart2018r2p2}
Rhinehart, N., Kitani, K.M., Vernaza, P.: R2p2: A reparameterized pushforward
  policy for diverse, precise generative path forecasting. In: Proceedings of
  the European Conference on Computer Vision (ECCV). pp. 772--788 (2018)

\bibitem{rissanen1996fisher}
Rissanen, J.J.: Fisher information and stochastic complexity. IEEE transactions
  on information theory  \textbf{42}(1),  40--47 (1996)

\bibitem{ruiz2018human}
Ruiz, A.H., Gall, J., Moreno-Noguer, F.: Human motion prediction via
  spatio-temporal inpainting. arXiv preprint arXiv:1812.05478  (2018)

\bibitem{seroussi1994algorithm}
Seroussi, B., Golmard, J.L.: An algorithm directly finding the k most probable
  configurations in bayesian networks. International Journal of Approximate
  Reasoning  \textbf{11}(3),  205--233 (1994)

\bibitem{sigal2010humaneva}
Sigal, L., Balan, A.O., Black, M.J.: Humaneva: Synchronized video and motion
  capture dataset and baseline algorithm for evaluation of articulated human
  motion. International journal of computer vision  \textbf{87}(1-2), ~4 (2010)

\bibitem{sourati2017probabilistic}
Sourati, J., Akcakaya, M., Erdogmus, D., Leen, T.K., Dy, J.G.: A probabilistic
  active learning algorithm based on fisher information ratio. IEEE
  transactions on pattern analysis and machine intelligence  \textbf{40}(8),
  2023--2029 (2017)

\bibitem{srivastava2017veegan}
Srivastava, A., Valkov, L., Russell, C., Gutmann, M.U., Sutton, C.: Veegan:
  Reducing mode collapse in gans using implicit variational learning. In:
  Advances in Neural Information Processing Systems. pp. 3308--3318 (2017)

\bibitem{tolstikhin2017wasserstein}
Tolstikhin, I., Bousquet, O., Gelly, S., Schoelkopf, B.: Wasserstein
  auto-encoders. arXiv preprint arXiv: 1711.01558  (2017)

\bibitem{troje2002decomposing}
Troje, N.F.: Decomposing biological motion: A framework for analysis and
  synthesis of human gait patterns. Journal of vision  \textbf{2}(5), ~2--2
  (2002)

\bibitem{walker2017pose}
Walker, J., Marino, K., Gupta, A., Hebert, M.: The pose knows: Video
  forecasting by generating pose futures. In: Proceedings of the IEEE
  international conference on computer vision. pp. 3332--3341 (2017)

\bibitem{wang2019imitation}
Wang, B., Adeli, E., Chiu, H.k., Huang, D.A., Niebles, J.C.: Imitation learning
  for human pose prediction. In: Proceedings of the IEEE International
  Conference on Computer Vision. pp. 7124--7133 (2019)

\bibitem{WengYuan2020}
Weng, X., Yuan, Y., Kitani, K.: Joint 3d tracking and forecasting with graph
  neural network and diversity sampling. arXiv:2003.07847  (2020)

\bibitem{yan2018mt}
Yan, X., Rastogi, A., Villegas, R., Sunkavalli, K., Shechtman, E., Hadap, S.,
  Yumer, E., Lee, H.: Mt-vae: Learning motion transformations to generate
  multimodal human dynamics. In: Proceedings of the European Conference on
  Computer Vision (ECCV). pp. 265--281 (2018)

\bibitem{yang2019diversity}
Yang, D., Hong, S., Jang, Y., Zhao, T., Lee, H.: Diversity-sensitive
  conditional generative adversarial networks. arXiv preprint arXiv:1901.09024
  (2019)

\bibitem{yang20183d}
Yang, W., Ouyang, W., Wang, X., Ren, J., Li, H., Wang, X.: 3d human pose
  estimation in the wild by adversarial learning. In: Proceedings of the IEEE
  Conference on Computer Vision and Pattern Recognition. pp. 5255--5264 (2018)

\bibitem{yuan2019diverse}
Yuan, Y., Kitani, K.: Diverse trajectory forecasting with determinantal point
  processes. arXiv preprint arXiv:1907.04967  (2019)

\bibitem{yuan2019egopose}
Yuan, Y., Kitani, K.: Ego-pose estimation and forecasting as real-time pd
  control. In: Proceedings of the IEEE International Conference on Computer
  Vision. pp. 10082--10092 (2019)

\bibitem{yuan2020residual}
Yuan, Y., Kitani, K.: Residual force control for agile human behavior imitation
  and extended motion synthesis. arXiv preprint arXiv:2006.07364  (2020)

\bibitem{zhang2019predicting}
Zhang, J.Y., Felsen, P., Kanazawa, A., Malik, J.: Predicting 3d human dynamics
  from video. In: Proceedings of the IEEE International Conference on Computer
  Vision. pp. 7114--7123 (2019)

\bibitem{zhao2017infovae}
Zhao, S., Song, J., Ermon, S.: Infovae: Information maximizing variational
  autoencoders. arXiv preprint arXiv:1706.02262  (2017)

\end{thebibliography}
	
	\appendix
	
\clearpage
\section{Additional Human3.6M Results}
\begin{figure}[ht!]
    \vspace{-7mm}
    \centering
    \includegraphics[width=0.95\textwidth]{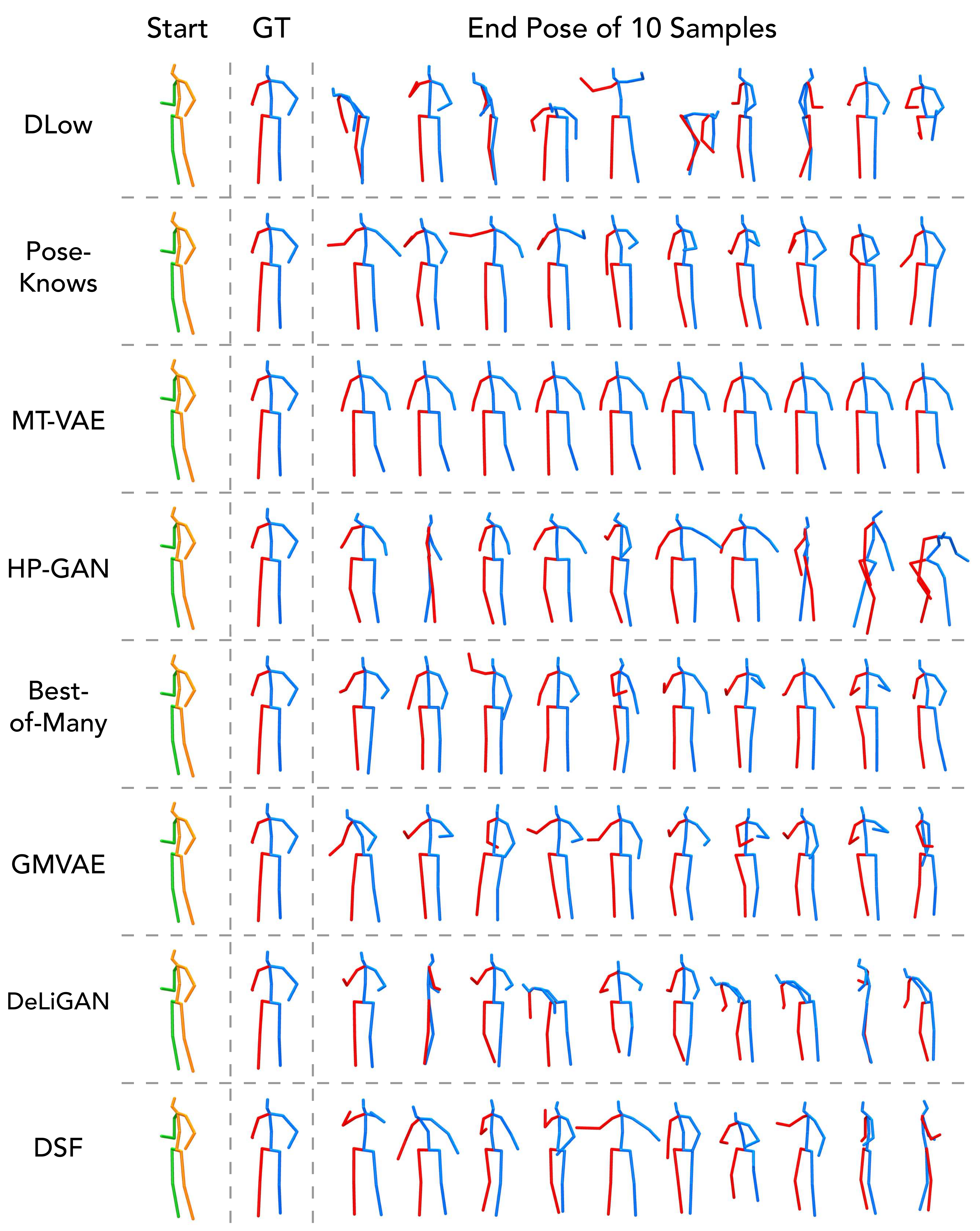}
    \vspace{-5mm}
    \caption{\textbf{Additional comparison with the baselines on Human3.6M.} We show the start pose, the end pose of the ground truth future motion, and the end pose of 10 motion samples by each method.}
    \label{fig:supp_h36m_comp}
\end{figure}

\begin{figure}[ht!]
    \centering
    \includegraphics[width=\textwidth]{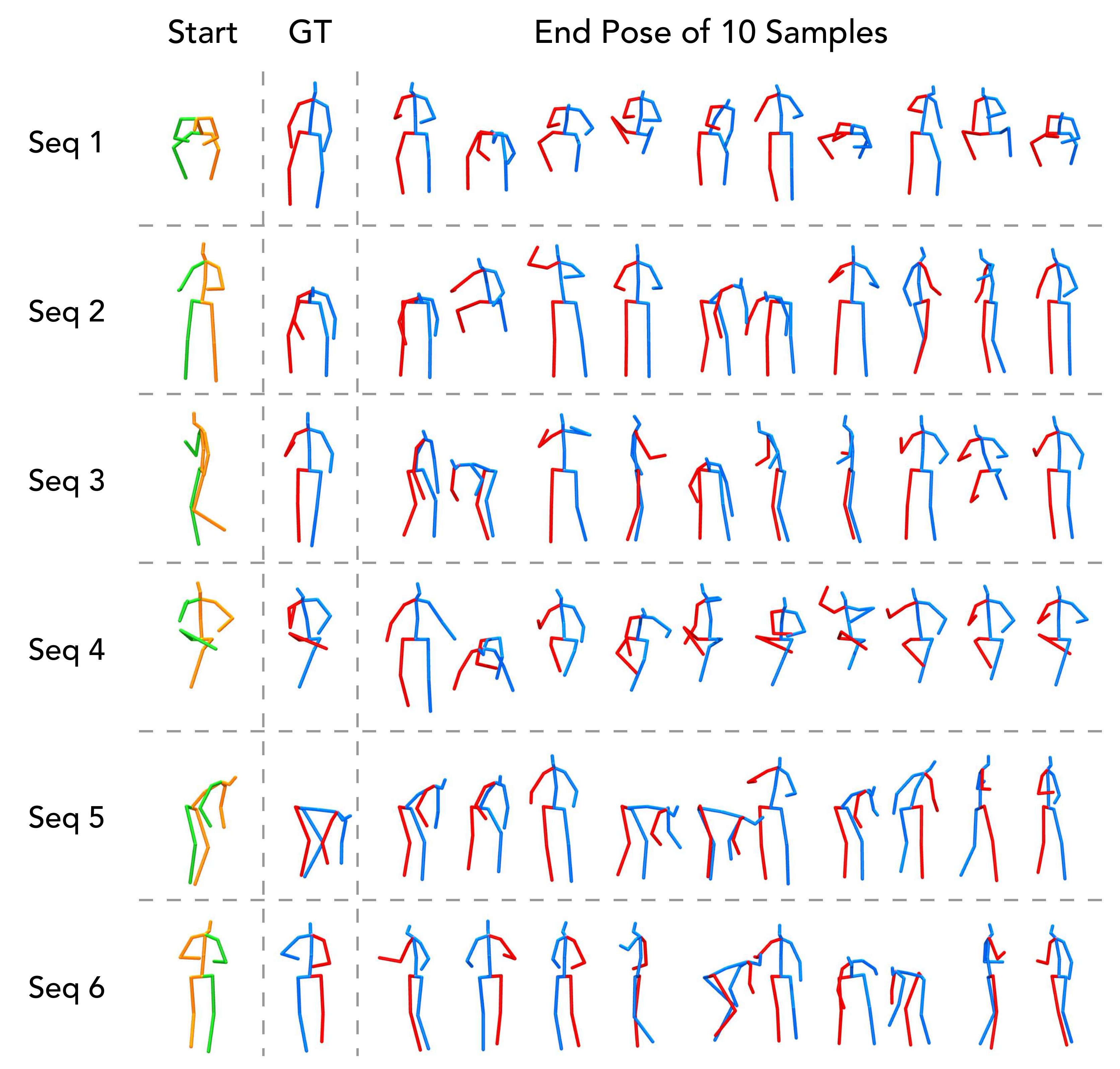}
    \caption{\textbf{Additional examples of DLow on Human3.6M.} Each row corresponds to a different sequence, where we show the start pose, the end pose of the ground truth future motion, and the end pose of 10 motion samples.}
    \label{fig:supp_h36m}
\end{figure}

In this section, we show more qualitative results on Human3.6M, including additional comparison with baselines (Fig.~\ref{fig:supp_h36m_comp}) and additional examples of DLow (Fig.~\ref{fig:supp_h36m}). Please refer to the \href{https://youtu.be/64OEdSadb00}{video} to see the whole motion sequences.
\vspace{-3mm}
\subsection{Additional Comparison with Baselines on Human3.6M}

\clearpage
\subsection{Additional Examples of DLow on Human3.6M}

\clearpage
\section{Additional Controllable Motion Prediction Results}
\label{sec:control_res}
\begin{figure}[ht!]
\vspace{-7mm}
    \centering
    \includegraphics[width=0.95\textwidth]{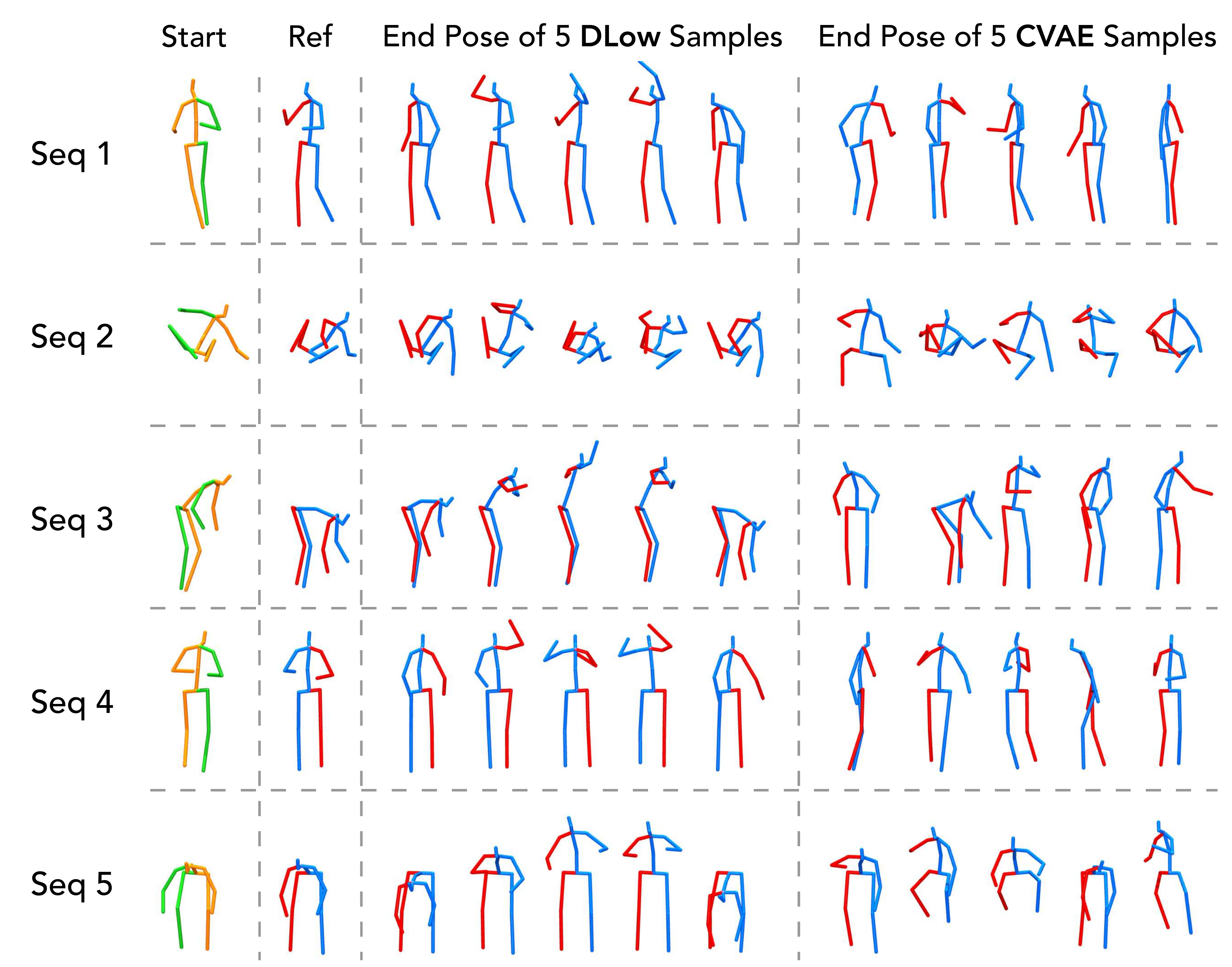}
    \vspace{-3mm}
    \caption{\textbf{Additional results on controllable motion prediction.} DLow can produce motion samples that have similar leg motion to the reference (Ref) yet diverse upper-body motion, while CVAE (random) samples cannot enforce similar leg motion.}
    \label{fig:control_supp}
    \vspace{-10mm}
\end{figure}

In Fig.~\ref{fig:control_supp}, we show additional results on controllable motion prediction using Human3.6M, where we use DLow to constrain the motion samples to have similar leg motion to the reference motion but diverse upper-body motion. Notice that DLow is able to produce samples with similar leg motion, while CVAE (random) samples cannot enforce similar leg motion. We further show some quantitative results in Table~\ref{table:control}, where we compute the average leg motion distance from motion samples to the reference motion and the APD for upper-body motion.

\vspace{1mm}\noindent\textbf{Implementation Details.}
We use the same networks in Fig.~3 of the main paper and the same hyperparmeters and training procedure given in the implementation details of the main paper. The main modification is that we use Eq.~24 in the paper for the energy function $E$ of the prior $p(X)$, and the DLow objective in Eq.~12 can be rewritten as: $L(\psi) = \beta L_\text{KL} + \lambda_d E_d + \lambda_s E_s + \lambda_r E_r$. We set $(\beta, \lambda_d, \lambda_s, \lambda_r)$ to $(1, 50, 10, 0)$. We also use a full parametrization of $\mathbf{A}_k$ instead of a diagonal one.

\begin{table}[ht!]
\footnotesize
\centering
\vspace{-5mm}
\begin{tabular}{c@{\hskip 2mm}c@{\hskip 2mm}c}
\toprule
Method & Leg Dist $\downarrow$ & Upper-body APD $\uparrow$ \\ \midrule
DLow & \textbf{1.071} & \textbf{12.741} \\  
CVAE & 2.958 & 6.051 \\  
\bottomrule
\end{tabular}
\vspace{2mm}
\caption{\textbf{Quantitative results} for controllable motion prediction.}
\label{table:control}
\vspace{-10mm}
\end{table}

\clearpage
\section{Metrics vs. Number of Samples $K$}
\label{sec:metrics_vs_k}
\begin{figure}[ht!]
    \vspace{-5mm}
    \centering
    \includegraphics[width=\textwidth]{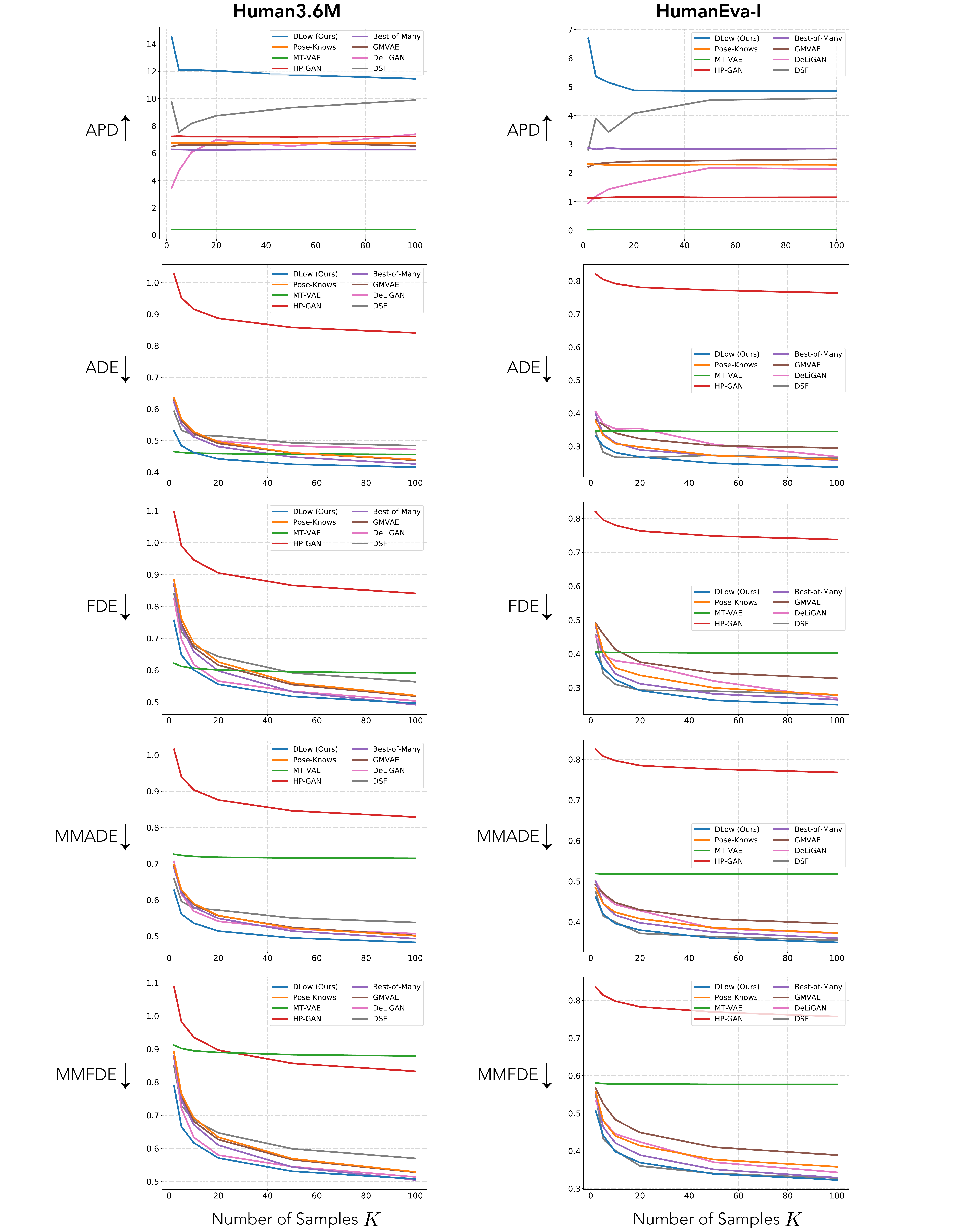}
    \caption{\textbf{Metrics vs. Number of Samples $K$} on both Human3.6M (Left) and HumanEva-I (Right).}
    \label{fig:metrics}
    \vspace{-5mm}
\end{figure}

\clearpage
\section{Additional HumanEva-I Results}
\begin{figure}[ht!]
    \vspace{-5mm}
    \centering
    \includegraphics[width=0.95\textwidth]{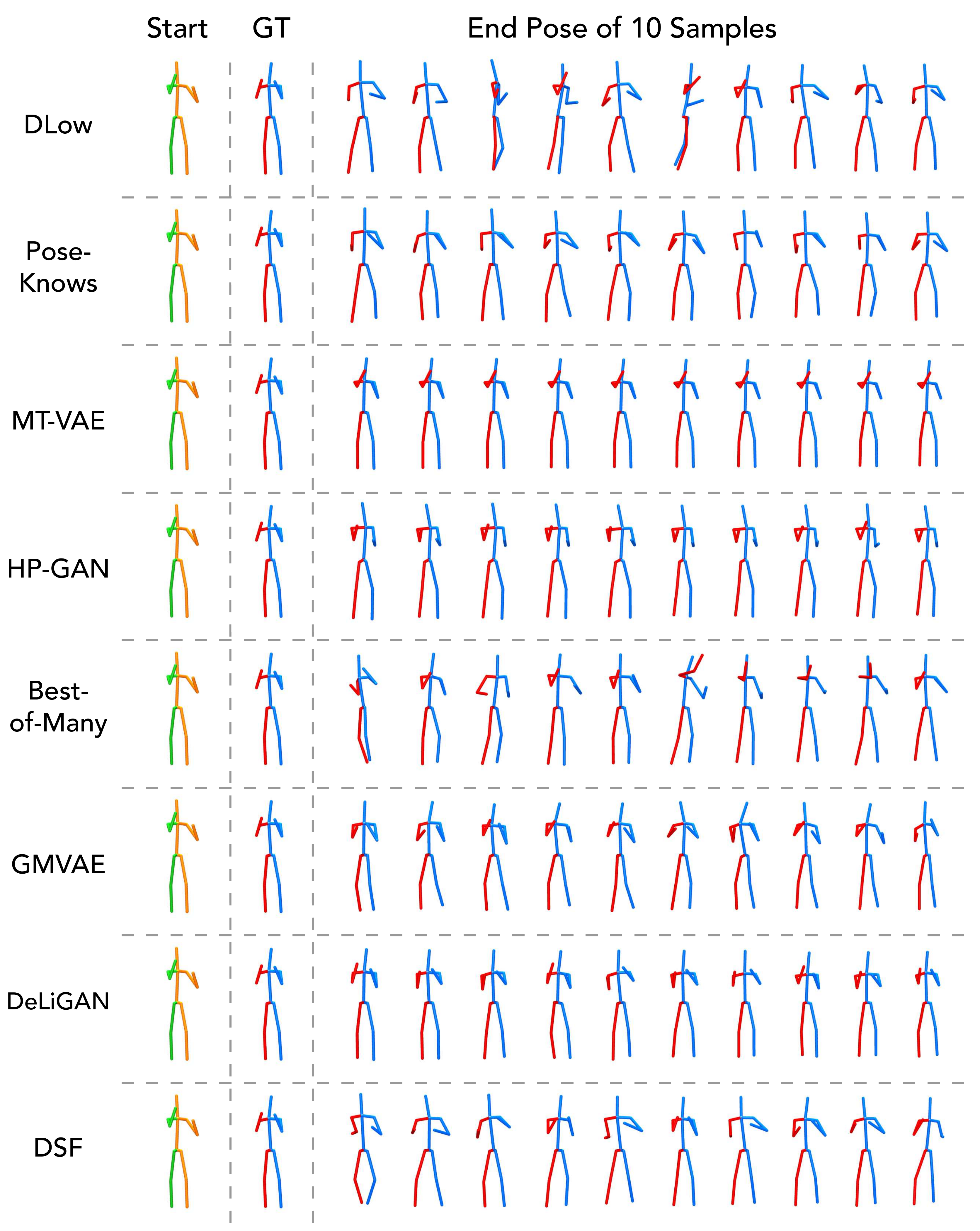}
    \vspace{-5mm}
    \caption{\textbf{Additional comparison with the baselines on HumanEva-I.} We show the start pose, the end pose of the ground truth future motion, and the end pose of 10 motion samples by each method.}
    \label{fig:supp_heva_comp}
    \vspace{-5mm}
\end{figure}

\begin{figure}[ht!]
    \centering
    \includegraphics[width=\textwidth]{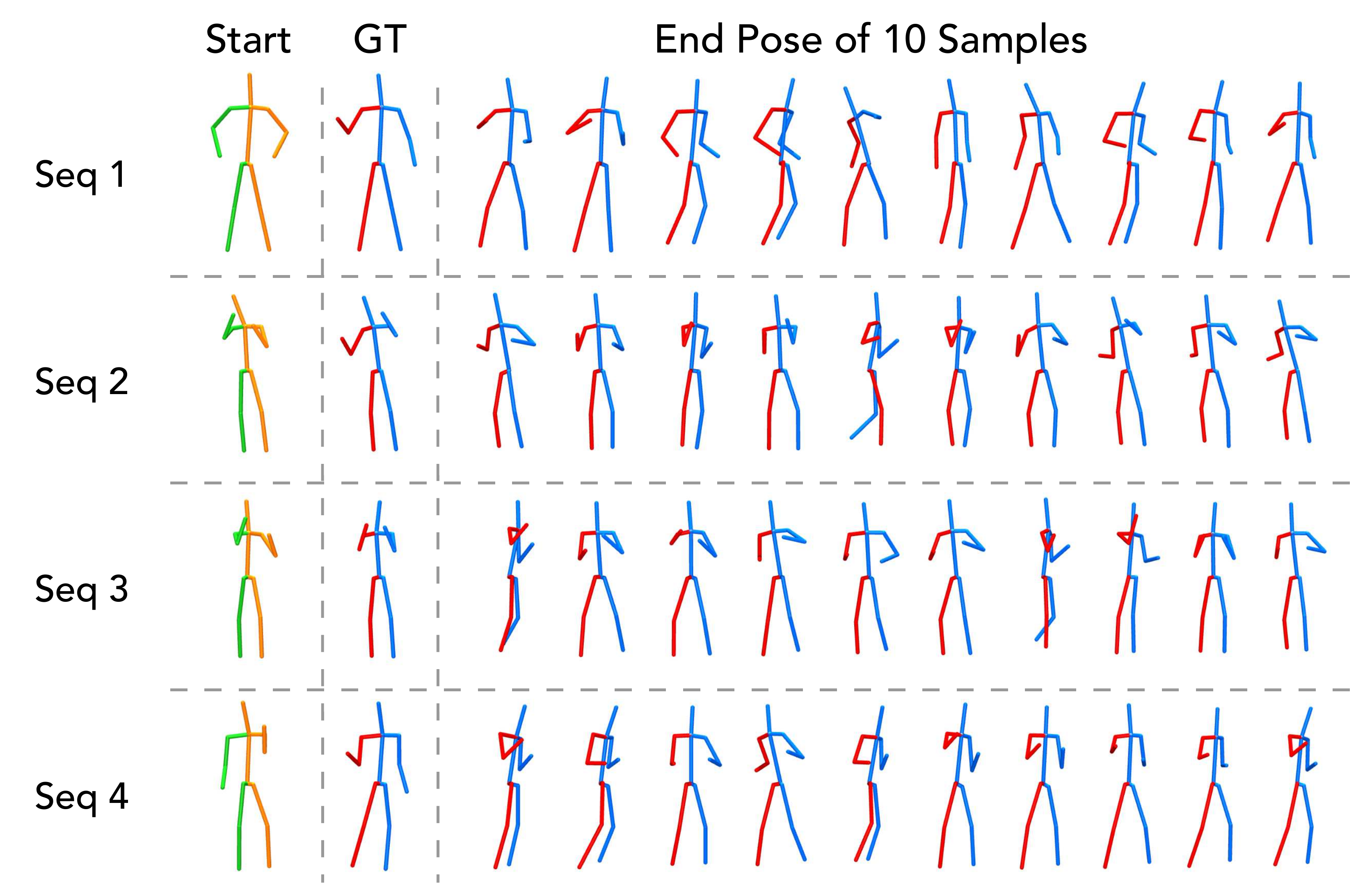}
    \caption{\textbf{Additional examples of DLow on HumanEva-I.} Each row corresponds to a different sequence, where we show the start pose, the end pose of the ground truth future motion, and the end pose of 10 motion samples.}
    \label{fig:supp_heva}
\end{figure}

We also show more qualitative results on HumanEva-I which is a much smaller dataset with less motion variation. We present additional comparison with baselines (Fig.~\ref{fig:supp_heva_comp}) and additional examples of DLow (Fig.~\ref{fig:supp_heva}).
\subsection{Additional Comparison with Baselines on HumanEva-I}

\clearpage
\subsection{Additional Examples of DLow on HumanEva-I}

\end{document}